\documentclass{article}

\usepackage[preprint]{neurips_2025}

\usepackage[utf8]{inputenc} 
\usepackage[T1]{fontenc}    
\usepackage{hyperref}       
\usepackage{url}            
\usepackage{booktabs}       
\usepackage{amsfonts}       
\usepackage{nicefrac}       
\usepackage{microtype}      
\usepackage{comment}
\usepackage{booktabs}
\usepackage{multicol}
\usepackage{multirow}
\usepackage{fontawesome}
\usepackage{tikz}
\usepackage{pifont}
\usepackage{wrapfig}
\usepackage{graphicx}
\usepackage{subcaption}
\usepackage{scalerel}
\usepackage{amssymb}
\usepackage{graphicx}
\usepackage{tablefootnote}
\usepackage{makecell}

\def\emojix{\scalerel*{\includegraphics{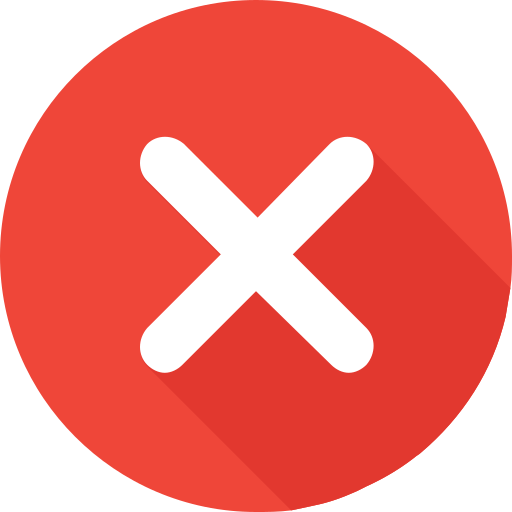}}{\small\textrm{\textbigcircle}}}
\def\emojij{\scalerel*{\includegraphics{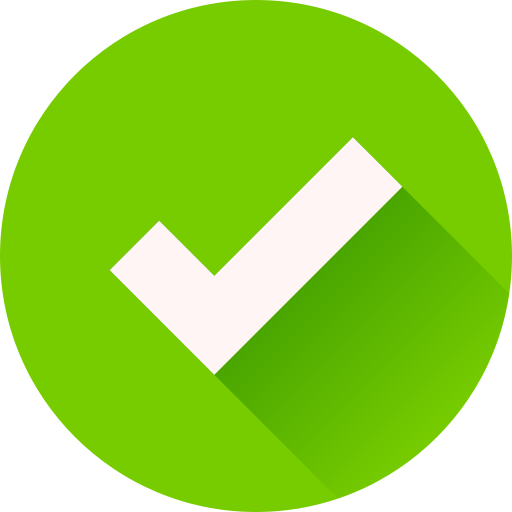}}{\small\textrm{\textbigcircle}}}

\definecolor{darkgreen}{RGB}{0,100,0}

\usepackage[many]{tcolorbox}
\usepackage{listings}
\usepackage{colortbl}
\usepackage[normalem]{ulem}
\useunder{\uline}{\ul}{}

\definecolor{promptblue}{RGB}{51,122,183}
\definecolor{inputgreen}{RGB}{60,118,61}
\definecolor{inputred}{RGB}{178,34,34}
\definecolor{systemgray}{RGB}{108,117,125}
\definecolor{positive}{RGB}{0,100,0}
\definecolor{negative}{RGB}{180,0,0}

\tcbset{
    promptbox/.style={
        colback=white,
        colframe=promptblue,
        boxrule=0.5pt,
        fontupper=\small\ttfamily,
        arc=2pt,
        boxsep=2pt,
        left=3pt,right=3pt,top=2pt,bottom=2pt,
        width=\columnwidth
    }
}

\newcommand{\dd}{{\textsc{DeepDialogue }}}
\newcommand{\ddnospace}{{\textsc{DeepDialogue}}}

\title{DeepDialogue: A Multi-Turn Emotionally-Rich Spoken Dialogue Dataset}

\author{%
  Alkis Koudounas \\
  Politecnico di Torino \\
  Turin, Italy \\
  \small{\texttt{alkis.koudounas@polito.it}} 
  \And
  Moreno La Quatra \\
  Kore University of Enna \\
  Enna, Italy \\
  \small{\texttt{moreno.laquatra@unikore.it}} 
  \And
  Elena Baralis \\
  Politecnico di Torino \\
  Turin, Italy \\
  \small{\texttt{elena.baralis@polito.it}}
}

\begin{document}

\maketitle

\begin{abstract}
Recent advances in conversational AI have demonstrated impressive capabilities in single-turn responses, yet multi-turn dialogues remain challenging for even the most sophisticated language models.
Current dialogue datasets are limited in their emotional range, domain diversity, turn depth, and are predominantly text-only, hindering progress in developing more human-like conversational systems across modalities.
To address these limitations, we present DeepDialogue, a large-scale multimodal dataset containing 40,150 high-quality multi-turn dialogues spanning 41 domains and incorporating 20 distinct emotions with coherent emotional progressions.
Our approach pairs 9 different language models (4B-72B parameters) to generate 65,600 initial conversations, which we then evaluate through a combination of human annotation and LLM-based quality filtering.
The resulting dataset reveals fundamental insights: smaller models fail to maintain coherence beyond 6 dialogue turns; concrete domains (e.g., ``cars,'' ``travel'') yield more meaningful conversations than abstract ones (e.g., ``philosophy''); and cross-model interactions produce more coherent dialogues than same-model conversations.
A key contribution of DeepDialogue is its speech component, where we synthesize emotion-consistent voices for all 40,150 dialogues, creating the first large-scale open-source multimodal dialogue dataset that faithfully preserves emotional context across multi-turn conversations. \texttt{\href{https://salt-research.github.io/DeepDialogue}{salt-research.github.io/DeepDialogue}}
\begin{figure}[!hb]
    \centering
    \includegraphics[width=\textwidth]{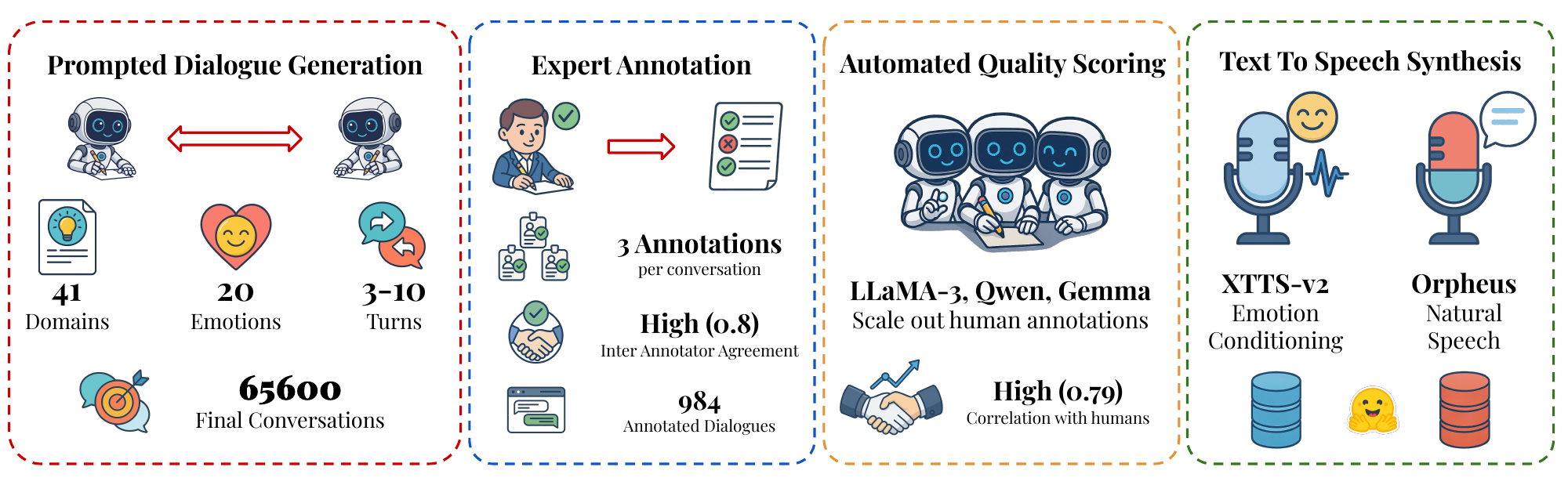}
    \caption{\dd dataset generation framework. The pipeline includes (1) text-based dialogue generation with emotion and domain conditioning, (2) human annotation, (3) automated LLM-based filtering, and (4) dual text-to-speech synthesis strategies.}
    \label{fig:dd-framework}
\end{figure}
\end{abstract}

\section{Introduction}
Conversational AI has witnessed impressive advancements in recent years, with large language models (LLMs) demonstrating unprecedented capabilities in understanding and generating human-like text.
These models excel at producing coherent single-turn responses across diverse topics and contexts \cite{brown2020language, openai2023gpt4}.
However, maintaining coherence, relevance, and emotional consistency across extended multi-turn dialogues remains a significant challenge even for state-of-the-art systems \cite{roller2021recipes, thoppilan2022lamda}.
This limitation becomes particularly evident when conversations extend beyond basic information exchange to include emotional nuances, domain-specific knowledge, and natural conversational flow.

Current dialogue datasets suffer from several limitations that slow down progress in multi-turn conversational AI research.
Existing resources typically lack emotional annotations at the turn level, provide insufficient domain coverage, or contain relatively short conversation sequences \cite{li2017dailydialog, si2023spokenwoz, chu2024towards}.
Additionally, most datasets are either human-human conversations that may not highlight the specific failure modes of AI systems, or human-AI interactions that reflect the limitations of the specific AI system used \cite{budzianowski2018multiwoz, zhang2018personalizing}.
The vast majority of available dialogue datasets are text-only, failing to capture the rich paralinguistic features that characterize human conversations \cite{poria2019emotion, wagner2023dawn}.
The scarcity of large-scale, emotionally-rich, multi-turn dialogue data spanning diverse domains and modalities creates a practical bottleneck for developing and evaluating more sophisticated conversational models.

To address these gaps, we introduce DeepDialogue, a comprehensive multimodal dataset containing 40,150 high-quality multi-turn conversations across 41 diverse domains, each incorporating emotional progression across 20 distinct emotional states.
Our methodology employs a systematic approach to dialogue generation by orchestrating interactions between 9 different text-only language models ranging from 4B to 72B parameters.
We design a careful prompting strategy that instructs models to maintain domain relevance while following specific ``emotional chains'' throughout conversations ranging from 3 to 10 turns.
This approach yields 65,600 initial dialogues, which undergo rigorous quality assessment through a combination of human evaluation (984 dialogues assessed by three independent annotators) and automated evaluation using a high-quality model judgment system that closely aligns with human preferences.
The resulting filtered dataset preserves only the highest quality conversations that maintain coherence, domain adherence, and emotional consistency.

We then transform all 40,150 text dialogues into spoken conversations using advanced text-to-speech (TTS) technology.
Unlike conventional TTS applications, our approach preserves the specific emotional context of each utterance, ensuring that the synthesized speech faithfully conveys the intended emotional states. 
This results in dialogues where emotions are expressed through both content and vocal characteristics, creating a more complete representation of conversational dynamics.
We specifically build two variants of our speech synthesis approach.
The first leverages the XTTS model \cite{xtts} that explicitly allows conditioning on reference audio samples from emotional speech datasets (e.g., RAVDESS~\cite{ravdess}).
This approach uses reference samples matching the required emotional state while maintaining actor consistency throughout dialogues, effectively simulating interactions between two distinct speakers with appropriate emotional expressions.
The second variant employs Orpheus \cite{orpheus} state-of-the-art TTS model, capable of generating highly natural speech but without explicit emotion conditioning mechanisms.
In this case, the model can only rely on linguistic cues like exclamations, word choice, and syntactic structures to generate appropriate prosodic variations related to the implicit emotion.
This dual approach allows researchers to compare explicitly-conditioned emotional speech with implicitly-inferred emotional expressions.
In both cases, for each dialogue, we assign consistent voice identities to each persona.

Our analysis of DeepDialogue reveals several important findings that advance our understanding of current LLM capabilities and limitations in dialogue generation.
In line with recent research~\cite{laban2025llmslostmultiturnconversation}, we observe a significant degradation in dialogue coherence and quality for smaller models (4-8B params) when conversations exceed 6 turns, with these models increasingly deviating from instructions, losing emotional consistency, and straying from the original domain.
In contrast, larger models (27-72B params) maintain relatively stable performance across longer conversation sequences, suggesting that parameter scaling remains important for handling extended dialogue contexts.

\begin{table}[!ht]
\caption{Stats of DeepDialogue and existing datasets. \texttt{*} means the information is not available.} 
\label{table-rw-datasets}
\centering
\scalebox{0.76}{%
\begin{tabular}{lcccccc}
\toprule
\textbf{Dataset}      
    & \textbf{\#Dialogues} 
    & \textbf{\#Turns} 
    & \textbf{Length (hrs)} 
    & \textbf{Audio} 
    & \textbf{\#Domains} 
    & \textbf{\#Emotions} \\
\midrule    
\texttt{KVRET}~\cite{eric2017key}
    & 2,425    
    & 12,732    
    & -    
    & \emojix
    & 3  
    & \emojix  \\
\texttt{M2M}~\cite{shah2018building}
    & 1,500     
    & 14,796    
    & -    
    & \emojix  
    & 2  
    & \emojix  \\
\texttt{MultiWOZ}~\cite{budzianowski2018multiwoz}
    & 8,438     
    & 113,556  
    & -    
    & \emojix  
    & 7  
    & \emojix  \\
\texttt{DailyDialog}~\cite{li2017dailydialog}
    & 13,118     
    & 103,632  
    & -    
    & \emojix  
    & 10  
    & 6  \\
\texttt{ABCD}~\cite{chen-etal-2021-action}
    & 8,034     
    & 177,407   
    & -    
    & \emojix        
    & 30 
    & \emojix  \\
\texttt{Chatbot Arena}~\cite{chiang2024chatbot}
    & 33,000    
    & 39,600    
    & -    
    & \emojix  
    & 8  
    & \emojix  \\
LMSYS-Chat-1M~\cite{zhenglmsys}
    & 1,000,000
    & 2,000,000
    & -    
    & \emojix  
    & \texttt{*} 
    & \emojix  \\
    \midrule
\texttt{IEMOCAP}~\cite{busso2008iemocap}
    & 151       
    & 10,039    
    & 12   
    & \emojij
    & \texttt{*} 
    & 4   \\
\texttt{DSTC2}~\cite{henderson-etal-2014-second}
    & 1,612     
    & 23,354    
    & 32   
    & \emojij
    & 1  
    & \emojix  \\
\texttt{DSTC10}~\cite{yoshino2023overview}
    & 107       
    & 2,292     
    & 45   
    & \emojij
    & 3  
    & \emojix  \\ 
\texttt{MELD}~\cite{poria2019meld}
    & 1,433     
    & 13,000    
    & 13.7 
    & \emojij
    & \texttt{*}  
    & 7   \\
\texttt{DailyTalk}~\cite{lee2023dailytalk}
    & 2,514     
    & 23,774    
    & 21.7 & \emojij
    & \texttt{*} 
    & \emojix  \\
\texttt{Expresso}~\cite{nguyen2023expresso}
    & 391       
    & 2,400     
    & 47   
    & \emojij
    & \texttt{*} 
    & 19 \\
\texttt{SpokenWOZ}~\cite{si2023spokenwoz}
    & 5,700     
    & 203,074   
    & 249  
    & \emojij
    & 7  
    & \emojix  \\
\texttt{MultiDialog}~\cite{park2024let}
    & 8,733     
    & 187,859   
    & 340  
    & \emojij
    & 8  
    & 8   \\
\texttt{SD-Eval (Emo)}~\cite{sdeval}
    & \texttt{*}         
    & 1,289     
    & 1,11 
    & \emojij
    & \texttt{*}  
    & 5   \\
\texttt{E-Chat200}~\cite{xue2024chat}
    & \texttt{*}         
    & 178,000   
    & 193  
    & \emojij
    & \texttt{*}  
    & 5   \\
\texttt{StyleTalk}~\cite{lin2024advancing}
    & 2,364     
    & 2,967     
    & \texttt{*}    
    & \emojij
    & 17 
    & 5   \\
\texttt{\textbf{DeepDialogue}}
    & \textbf{40,150}  
    & \textbf{241,825}
    & \textbf{488}          
    & \emojij           
    & \textbf{41}        
    & \textbf{20 (text)}    \\
\bottomrule
\end{tabular}}
\end{table}
Domain type emerges as another critical factor affecting dialogue quality across all model sizes.
Our analysis supports previous findings~\cite{liao2023concept}, revealing that concrete domains with clearly defined boundaries (e.g., ``cars,'' ``travel,'' ``cooking'') consistently yield more grounded, specific, and meaningful conversations compared to abstract domains (e.g., ``philosophy,'' ``spirituality'').
In concrete domains, models more frequently reference specific entities, share plausible experiences, and reach natural conversation conclusions.
Conversely, dialogues in abstract domains often exhibit circular reasoning, vague statements, and fail to reach substantive conclusions.

Finally, we notice that dialogues generated by pairing different models consistently demonstrate higher quality metrics compared to those where the same model converses with itself.
This ``cross-model effect'' holds true across model sizes and domains, with mixed-model conversations exhibiting greater diversity in response patterns, more natural turn-taking dynamics, and fewer instances of repetitive or circular exchanges. 
We hypothesize this effect comes from the varied training distributions and inductive biases of different models, creating complementary strengths that enhance conversational dynamics.
When a single model converses with itself, it may amplify its own patterns and limitations, whereas different models can compensate for each other's weaknesses, similar to diverse perspectives in human conversations.

The contributions of this work are as follows: 
(1) We release DeepDialogue, a large-scale multimodal dataset of LLM-generated multi-turn dialogues with explicit emotional annotations and transitions, enabling research on emotionally intelligent conversational systems;
(2) DeepDialogue spans 41 distinct domains, 14 different model combinations, and 7 turn types, allowing for a comprehensive analysis of domain- and model-specific dialogue patterns and challenges;
(3) DeepDialogue offers both text and emotionally-appropriate speech representations for all dialogues, creating a resource that bridges the gap between text-only dialogue research and real-world spoken conversation applications.

\section{Related Works}
\label{section:rw}
\textbf{Text-based Dialogues.}
The development of dialogue systems has been largely driven by written dialogue datasets. Early corpora such as KVRET~\cite{eric2017key} and M2M~\cite{shah2018building} focused on human-machine interactions within very few domain settings. Subsequently, datasets like MultiWOZ~\cite{budzianowski2018multiwoz}, DailyDialog~\cite{li2017dailydialog}, and ABCD~\cite{chen-etal-2021-action} expanded to multi-domain dialogues, aiming to better approximate real-world scenarios. In parallel, large-scale evaluation benchmarks like Chatbot Arena~\cite{chiang2024chatbot} and LMSYS-Chat-1M~\cite{zhenglmsys} have emerged to assess general-purpose conversational agents. While these datasets provide valuable insights into model capabilities and human preferences, they are not task-specific, lack domain annotations, and do not include audio or emotional context, thus limiting their utility for developing grounded spoken dialogue systems.

\textbf{Spoken and Multimodal Dialogues.}
Spoken research has been supported by datasets such as DSTC2~\cite{henderson-etal-2014-second}, though these remain limited in scale and domain diversity. DSTC10~\cite{yoshino2023overview} revisits this space by adding ASR hypotheses, while SpokenWOZ~\cite{si2023spokenwoz} introduces a large-scale speech-text dataset with full annotations. Further progress includes DailyTalk~\cite{lee2023dailytalk}, which presents natural speech conversations, and MultiDialog~\cite{park2024let}, which extends this to multimodal settings with audio-visual content. Still, many spoken datasets remain constrained by scale, domain scope, or lack fine-grained style and emotion information.

\textbf{Speaking Style and Emotion in Dialogues.}
Understanding speaking style is critical for generating human-like spoken dialogue. Datasets such as IEMOCAP~\cite{busso2008iemocap}, RAVDESS~\cite{ravdess} and MELD~\cite{poria2019meld} have been widely used to study emotional dynamics, typically using acted or scripted dialogues. Others like Expresso~\cite{nguyen2023expresso} and StyleTalk~\cite{lin2024advancing} introduce stylistically diverse responses for more expressive modeling. However, these datasets are typically not aligned with task-oriented dialogue settings, limiting their applicability for modeling goal-directed conversations that also require nuanced emotional or stylistic variation.
Additional datasets such as SD-Eval (Emo)~\cite{sdeval} and E-Chat200~\cite{xue2024chat} have emerged to support emotion-aware evaluation and stylized dialogue modeling using speech data. Yet, they often lack comprehensive domain coverage or conversational structure, which restricts their use in building holistic, multi-domain dialogue systems.

\textbf{Positioning Our Work.}
In contrast to existing datasets that typically emphasize one aspect, be it conversation, emotion, or style, DeepDialogue introduces a large-scale speech-text corpus that spans 41 domains and includes 20 emotion categories (Table~\ref{table-rw-datasets}). It offers a unified benchmark for studying style-sensitive spoken dialogues, grounded in diverse, multi-domain, emotionally expressive conversations. This enables more comprehensive modeling of naturalistic dialogue systems, where task, style, and affect interact.

\section{DeepDialogue}~\label{section:methodology}
We design a multi-stage pipeline for generating emotionally coherent, multi-turn dialogues across diverse domains using LLMs.
Our methodology consists of four main components:
(1) domain and emotion design;
(2) prompt-based dialogue generation and emotional progression modeling;
(3) quality evaluation and filtering;
(4) speech synthesis.

\subsection{Domain and Emotion Setup}
We define 41 conversation domains that range from concrete (e.g., \textit{travel}, \textit{cars}, \textit{cooking}) to abstract (e.g., \textit{philosophy}, \textit{spirituality}, \textit{politics}).
This diverse domain selection enables systematic investigation of how topic concreteness affects dialogue quality and coherence across different model sizes.
Concrete domains generally contain well-defined entities, activities, and knowledge structures that provide clear anchoring points for conversation, while abstract domains involve conceptual reasoning and value judgments that may challenge LLM capabilities differently.
The domain distribution was carefully balanced to represent everyday conversational topics (e.g., \textit{weather}, \textit{work}), specialized knowledge areas (e.g., \textit{science}, \textit{finance}), and social-emotional contexts (e.g., \textit{relationships}, \textit{health}).
For each domain, we associate a set of relevant emotions selected to reflect typical affective expressions in that context.
A total of 20 distinct emotions are used throughout the dataset, grouped according to semantic and affective similarity.
These emotion categories include basic emotions (e.g., \textit{happiness}, \textit{sadness}, \textit{anger}), complex social emotions (e.g., \textit{pride}, \textit{embarrassment}, \textit{gratitude}), and epistemic emotions (e.g., \textit{curiosity}, \textit{confusion}, \textit{surprise}).
The emotion-domain pairings were determined through a pilot study in which human annotators rated the naturalness of expressing specific emotions within each domain context, ensuring that the dataset reflects plausible emotional dynamics for real-world conversations.

We also define a structured mapping of emotion transitions, representing plausible progressions of emotional states across dialogue turns.
These mappings ensure that conversations evolve with emotionally realistic dynamics rather than exhibiting implausible emotional shifts.
For example, transitions from \textit{frustration} might lead to \textit{anger}, \textit{disappointment}, or \textit{anxiety}, but would rarely jump directly to \textit{excitement} or \textit{happiness} without intermediate states.
We developed a directed graph of permissible emotion transitions based on psychological literature on emotional progression \cite{russell1980circumplex, plutchik2001nature}.
This approach allows us to construct ``emotional arcs'' for dialogues of different lengths, incorporating both predictable patterns (e.g., \textit{curiosity} $\rightarrow$ \textit{surprise} $\rightarrow$ \textit{excitement}) and more complex trajectories that reflect the nuanced nature of human emotional expression.
The resulting emotion transition framework provides a structured yet flexible mechanism for generating dialogues with coherent emotional progression while maintaining diversity in conversational dynamics.
More details on domains, emotions, and mappings are described in the Appendix.

\subsection{Dialogue Generation Process and Emotion Progression Modeling}
To create each dialogue, we begin by randomly selecting a domain and sampling an initial emotion relevant to that domain.
This selection process follows a stratified sampling approach to ensure a balanced representation across the 41 domains and 20 emotions while respecting the domain-specific emotion constraints established in our framework.
We then initiate a conversation between two LLM agents, which take turns responding to each other.
For agent selection, we pair models from a collection of 9 text-only instruction-tuned LLMs\footnote{We employ the following models: \texttt{Llama-3.1-8B-Instruct}~\cite{grattafiori2024llama}, \texttt{Llama-3.3-70B-Instruct}~\cite{grattafiori2024llama}, \texttt{Qwen2.5-32B-Instruct}~\cite{yang2024qwen2}, \texttt{Qwen2.5-72B-Instruct}~\cite{yang2024qwen2},  \texttt{phi4-mini-instruct}~\cite{abouelenin2025phi}, \texttt{phi-4}~\cite{abdin2024phi}, \texttt{c4ai-command-r7b}~\cite{cohere2025command}, \texttt{gemma3-4B-Instruct}~\cite{team2025gemma}, \texttt{gemma3-27B-Instruct}~\cite{team2025gemma}.}, ranging from 4B to 72B parameters. We obtain 14 different pairs that include both same-model pairings and cross-model interactions.
Each model generates responses informed by conversation history, domain context, and its current emotional state.
We implement this through prompting techniques that provide the model with three key components: (1) the full conversation history, (2) explicit instructions about the domain of conversation, and (3) the specific emotion to express in the current turn.
To ensure high-quality outputs, we employ few-shot examples demonstrating appropriate emotional expression within similar domains, and we enforce a maximum response length of 25 words to maintain a natural conversational flow.

At every turn, the next emotion is determined using the predefined emotional progression rules, with additional consideration of domain appropriateness and conversational tone.
This dynamic emotion selection draws from the permitted transitions in our emotional graph structure, weighted by transition probabilities that favor contextually appropriate progressions.
For instance, in a \textit{travel} domain dialogue where one participant expresses \textit{excitement} about an upcoming trip, the response emotion might be sampled from weighted options including \textit{enthusiasm}, \textit{curiosity}, or \textit{amusement}, reflecting natural human reactions. 

Such a probabilistic approach introduces variability while maintaining emotional coherence throughout the conversation.
This content-aware emotional routing is further refined by analyzing the specific textual content of previous messages.
A lightweight sentiment analyzer identifies significant emotional cues in the previous turn, and when such cues are detected, the transition probabilities are dynamically adjusted.
For example, if a message contains an expression of personal achievement, the probability of transitioning to emotions like \textit{pride}, \textit{curiosity}, or \textit{excitement} increases, while transitions to \textit{boredom} or \textit{frustration} are suppressed.

We implement additional constraints that modulate transition probabilities based on conversation length.
Shorter dialogues (3-4 turns) exhibit more focused emotional arcs, while longer conversations (7-10 turns) explore more varied emotional territory with gradual transitions between affective states.
Each dialogue continues for a randomly chosen number of turns, between three and ten, to simulate natural dialogue variation. 

All responses are designed to resemble friendly, emotionally expressive exchanges between humans, avoiding system-level phrasing or rigid structures.
To achieve this, we explicitly instruct models to adopt a conversational persona and express emotions through both content and linguistic style (e.g., using exclamation marks for excitement, ellipses for hesitation, or descriptive language for emotional states). Examples of model prompts are shown in the Appendix. 

\begin{table}[!ht]
      \caption{\textbf{Agreement with human annotations.} Cohen Kappa (on decision score), Kendall Tau and Spearman correlation (on goodness score) between human and LLM annotations. Best results per category (closed-source, open-source, ensemble) underlined, best results overall in bold. Model used for final filtering highlighted in \colorbox[HTML]{DAE8FC}{light-blue}. We also report the percentage of accepted dialogues.} 
      \label{table-agreement-human}
      \centering
    \scalebox{0.76}{%
    \begin{tabular}{clcccc}
    \toprule
    \begin{tabular}[c]{@{}c@{}}\textbf{Model} \\ \textbf{Type}\end{tabular}
        & \multicolumn{1}{l}{\textbf{Model Name}} 
        & \begin{tabular}[c]{@{}c@{}}\textbf{Cohen Kappa} \\ \textbf{(Decision)}\end{tabular}
        & \begin{tabular}[c]{@{}c@{}}\textbf{Kendall-Tau} \\ \textbf{(Goodness)}\end{tabular}
        & \begin{tabular}[c]{@{}c@{}}\textbf{Spearman} \\ \textbf{(Goodness)}\end{tabular} 
        & \begin{tabular}[c]{@{}c@{}}\textbf{\% Good Dialogues} \\ \textbf{ (GT: 59\%)}\end{tabular} \\
    \midrule
    
    &
        \texttt{GPT-3.5-Turbo} 
        & 0.47 
        & 0.37 
        & 0.51 
        & 76\% \\
      
    &
        \texttt{GPT-4o} 
        & {\ul0.78} 
        & \textbf{{\ul0.64}} 
        & \textbf{{\ul0.79}} 
        & 54\% \\
      
    &
        \texttt{GPT-4o-mini} 
        & 0.64 
        & 0.52 
        & 0.66 
        & 73\% \\
    
    &
        \texttt{Gemini-2.0-Flash} 
        & 0.61 
        & 0.58 
        & 0.70
        & 49\% \\
    
    &
        \texttt{Gemini-2.5-Flash} 
        & 0.65 
        & 0.62 
        & 0.76 
        & 46\% \\
        
    \parbox[t]{2mm}{\multirow{-6}{*}{\rotatebox[origin=c]{90}{\textbf{Closed}}}}
    &
        \texttt{Gemini-2.5-Pro} 
        & 0.56 
        & 0.60 
        & 0.77 
        & 65\% \\

    \midrule

    &
        \texttt{Phi4} 
        & 0.39 
        & 0.49 
        & 0.59 
        & 85\% \\
        
    &
        \texttt{Gemma3-27B-it} 
        & 0.68 
        & {\ul 0.62} 
        & {\ul 0.75} 
        & 51\% \\
    
    &
        \texttt{Llama-3.3-70B-it} 
        & 0.72 
        & 0.60
        & 0.71 
        & 65\% \\
        
    &
        \texttt{Qwen2.5-72B-it} 
        & {\ul 0.76} 
        & 0.55 
        & 0.70 
        & 63\% \\

    \parbox[t]{2mm}{\multirow{-5}{*}{\rotatebox[origin=c]{90}{\textbf{Open}}}}
    &
        \texttt{Qwen1.5-110B-Chat} 
        & 0.35 
        & 0.54 
        & 0.64 
        & 86\% \\

    \midrule
    
    &
        \begin{tabular}[l]{@{}l@{}}\texttt{Llama-3.3-70B-it +} \\ \texttt{Qwen2.5-72B-it + Phi-4}\end{tabular}
        & 0.71 
        & 0.56 
        & 0.75 
        & 69\% \\
        
    &
        \begin{tabular}[l]{@{}l@{}}\texttt{Llama-3.3-70B-it +} \\ \texttt{Qwen2.5-72B-it + Qwen1.5-110B-Chat}\end{tabular}
        & 0.70
        & 0.57
        & 0.75 
        & 69\% \\

    \parbox[t]{2mm}{\multirow{-6}{*}{\rotatebox[origin=c]{90}{\textbf{Ensemble}}}}
    &
        \cellcolor[HTML]{DAE8FC}\begin{tabular}[l]{@{}l@{}}\texttt{Llama-3.3-70B-it +} \\ \texttt{Qwen2.5-72B-it + Gemma3-27B-it}\end{tabular}
        & \cellcolor[HTML]{DAE8FC}\textbf{{\ul 0.79}}
        & \cellcolor[HTML]{DAE8FC}{\ul 0.62} 
        & \cellcolor[HTML]{DAE8FC}\textbf{{\ul 0.79}}
        & \cellcolor[HTML]{DAE8FC}60\% \\
     
        

    \bottomrule
    
    \end{tabular}}
\end{table}
The process is repeated across all domains, generating 1,600 dialogues per domain for a total of 65,600 raw dialogues.
The resulting corpus contains diverse conversation dynamics, with systematic variation across domain types, emotional progressions, model pairings, and conversation lengths, enabling comprehensive analysis of these factors on dialogue quality and coherence.
More details on the specific model pairs used and dataset statistics can be found in the Appendix. 

\subsection{Quality Evaluation and Filtering}
Following generation, we evaluate the raw dialogues using a hybrid human-AI feedback approach.
First, a subset of 984 dialogues (24 per domain, 123 per number of turns) is rated by three human annotators based on dialogue coherence, emotional consistency, and domain relevance.
Each annotator is asked to provide a score on a Likert-scale from 1 to 5 for the overall quality of the dialogue. They also provide a final binary score (1 if they consider the dialogue to be valid, 0 otherwise).
Annotators must provide specific justifications when assigning negative scores, identifying whether issues stem from incoherence, emotional inconsistency, domain drift, model hallucinations and out-of-the-blue answers, or other problems.
We calculate inter-annotator agreement using Fleiss' kappa \cite{fleiss1973equivalence}, achieving a value of $\kappa = 0.80$, indicating substantial agreement.
These human annotations serve as reference data for calibrating an automatic filtering model.
We then benchmark several state-of-the-art LLMs to identify which most closely approximates human quality judgments.
Our evaluation includes open-source models (\texttt{Phi-4}~\cite{abdin2024phi}, \texttt{Gemma-3-27B-Instruct}~\cite{team2025gemma}, \texttt{Llama-3.3-70B-Instruct}~\cite{grattafiori2024llama}, \texttt{Qwen2.5-72B-Instruct}~\cite{yang2024qwen2}, \texttt{Qwen1.5-110B-Chat}~\cite{bai2023qwen}) and closed-source models (\texttt{GPT-3.5 Turbo}~\cite{brown2020language}, \texttt{GPT-4o-mini}~\cite{gpt4o-mini}, \texttt{GPT-4o}~\cite{hurst2024gpt}, \texttt{Gemini 2.0 Flash}~\cite{gemini2flash}, \texttt{Gemini 2.5 Flash}~\cite{gemini25flash}, \texttt{Gemini 2.5 Pro}~\cite{gemini25pro}).
Each model evaluates the human-annotated subset using identical prompts that detail the assessment criteria. Specifically, following human annotations, we ask the model to assess each dialogue across three dimensions: (1) Coherence: logical flow, natural turn transitions, and absence of contradictions, (2) Emotional consistency: adherence to the specified emotional progression, and (3) Domain relevance: maintenance of the assigned conversational domain.
The model assigns a final overall score (1-5)  and provides a final binary validity judgment.

We calculate agreement scores between each LLM's ratings and the human annotations, measuring both Kendall Tau and Spearman correlation for Likert scores and Cohen Kappa scores for binary decisions.
Results are shown in Table~\ref{table-agreement-human}. GPT-4o achieves the strongest alignment with human judgment, achieving a Cohen Kappa score $r = 0.78$ and a Kendall Tau score $\tau = 0.64$. 
We also experiment with a majority-voting ensemble of different configurations of open-source models, and demonstrate that combining \texttt{Gemma-3-27B-Instruct}, \texttt{Llama-3.3-70B-Instruct}, and \texttt{Qwen2.5-72B-Instruct} achieves superior or on-par performance with GPT-4o across all metrics (Cohen Kappa $r = 0.79$, Kendall Tau $\tau = 0.62$, Spearman $\rho = 0.79$).

Based on these results, we employ such an ensemble of open-source models to evaluate the full dataset of 65,600 initial dialogues.
We follow the annotation strategy discussed above and implement additional quality controls, including random spot checks by human reviewers to ensure the LLM evaluations maintain consistency throughout the process.
Only dialogues meeting minimum quality thresholds (scores $\geq 3$ on all dimensions and a positive binary judgment) are retained.
To ensure the absence of harmful, toxic, or unsafe content, we incorporate Llama Guard~\cite{grattafiori2024llama}, a safety-focused LLM classifier, into our pipeline. Each dialogue is passed through this model, and only those that pass its safety filters are considered for inclusion in the final dataset.

This whole filtering process yields 40,150 high-quality dialogues (approximately 61\% of the initial generation) for inclusion in the final DeepDialogue dataset. 
The final distribution of conversation lengths has a mean of 6.1 turns.
The filtering effectively removes conversations exhibiting hallucinations, logical contradictions, unnatural topic shifts, emotional incoherence, or domain drift.
Our approach demonstrates that carefully selected LLMs can effectively emulate human quality judgments for dialogue evaluation at scale, offering significant advantages in cost and time efficiency compared to full human annotation.
The resulting dataset represents a curated collection of dialogues that maintain coherence, emotional progression, and topical focus across varying conversation lengths and domains.

\subsection{Speech Synthesis}
To expand beyond text-only conversations, we synthesize emotionally expressive speech for all 40,150 dialogues.
Each turn is converted into audio using two different text-to-speech approaches, creating parallel versions of our dataset with distinct synthesis strategies.

The first approach uses \texttt{XTTS-v2}~\cite{xtts}, a model that enables zero-shot reference voice conditioning\footnote{\url{https://huggingface.co/datasets/SALT-Research/DeepDialogue-xtts}}.
We leverage the RAVDESS dataset~\cite{ravdess}, which contains recordings of 24 actors expressing 8 emotional states (7 emotions plus neutral).
To apply our more granular emotional taxonomy, we map our 20 emotional states to RAVDESS's 8 emotions; for example, \textit{``excited,''} \textit{``amused,''} and \textit{``enthusiastic''} are mapped to \textit{``happy,''} while \textit{``anxious''} maps to \textit{``fearful''} and \textit{``disappointed''} to \textit{``disgust.''} 
For each actor-emotion combination, we concatenate the two standard RAVDESS sentences to create reference audio samples, using only normal intensity recordings rather than strong intensity to avoid overemphasized emotional expressions.
This integration of emotional reference samples directly into the synthesis process produces natural vocalizations that convey subtle emotional variations throughout the conversation.

Our second variant employs \texttt{Orpheus}~\cite{orpheus}, a state-of-the-art TTS model that produces exceptionally natural speech but cannot be explicitly conditioned on emotional reference samples\footnote{\url{https://huggingface.co/datasets/SALT-Research/DeepDialogue-orpheus}}.
Orpheus offers 8 distinct voices, which we assign consistently to speakers throughout each dialogue.
With this model, emotional expressions must be inferred directly from the text itself, using linguistic cues such as exclamations, word choice, and sentence structure to generate appropriate emotional variations.

For both approaches, we assign a consistent voice identity to each conversational agent throughout the dialogue, with voices randomly selected from RAVDESS actors for \texttt{XTTS-v2} or from the pre-defined voice options for \texttt{Orpheus}.
We also preprocess text by removing non-speech symbols and emojis while preserving punctuation, which is crucial for conveying emotion.
The result is a comprehensive multimodal dataset where both text and audio consistently reflect the intended emotional progressions across extended dialogues.
The speech component for each variant contains utterances from all 40,150 conversations, totaling over 480 hours of audio across more than 240,000 individual turns for each variant.
This comprehensive corpus of emotionally varied speech spans our complete dataset, providing unprecedented scale for spoken dialogue research.
This multimodal extension enables novel research in dialogue understanding, speech-language model integration, and emotion-aware spoken systems.
The audio component of DeepDialogue aims at addressing the critical gap in current conversational AI resources, bridging text-based dialogue research with real-world spoken conversation applications.

\begin{figure*}
    \centering
    \begin{subfigure}[t]{0.40\textwidth}
        \centering
        \includegraphics[width=\linewidth,height=50mm, keepaspectratio]{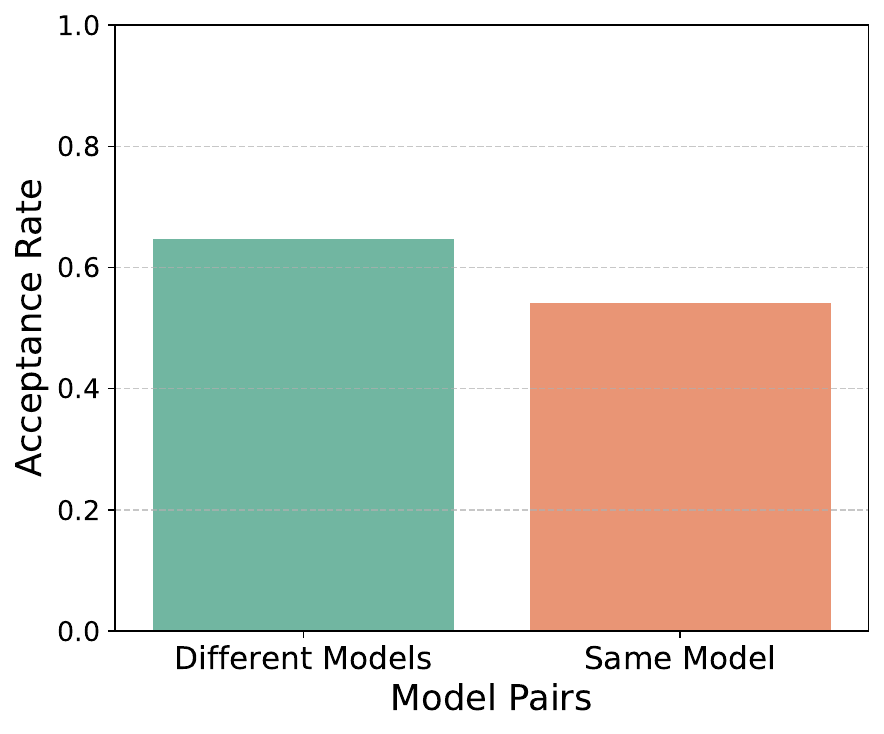}
        \caption{\textit{AR same vs. different model pairs}}
        \label{fig:model-pairs-a}
    \end{subfigure}
    \hfill
    \begin{subfigure}[t]{0.55\textwidth}
        \centering
        \includegraphics[width=\linewidth,height=55mm, keepaspectratio]{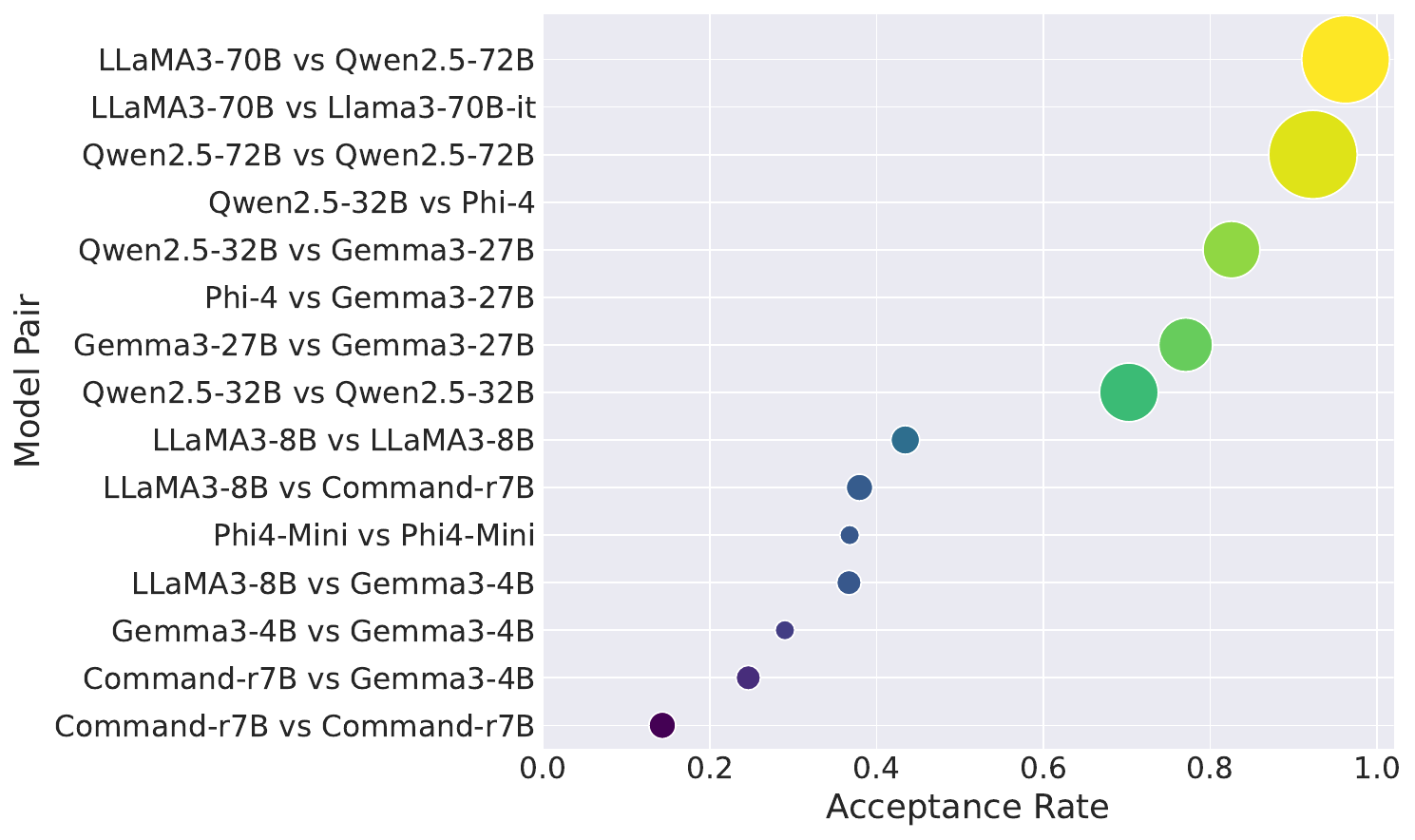}
        \caption{\textit{AR by model pairs}}
        \label{fig:model-pairs-b}
    \end{subfigure}
    \caption{Acceptance rate (AR) for pairs considering same vs. different models.}

    \label{fig:model-pairs-ab}
\end{figure*}

\section{Results and Discussion}~\label{section:analysis}
In this section, we first conduct an in-depth analysis of DeepDialogue's properties and then demonstrate the quality of its emotional content by using it to train models for speech emotion recognition.

\subsection{Empirical Findings from Human Annotations}

We consider here the subset of dialogues for which we collected human annotations. Further analysis on the whole dataset can be found in the Appendix.

\textbf{Cross-model effect}
Figure~\ref{fig:model-pairs-ab} illustrates the impact of model pairing strategies on dialogue acceptance rates.
Specifically, Figure~\ref{fig:model-pairs-a} compares the overall acceptance rate for dialogues generated by pairs of different LLMs versus dialogues where the same model converses with itself. This figure clearly demonstrates that pairing different models results in a higher acceptance rate (0.65) compared to using identical models for both conversational agents (0.54). This supports our finding of a beneficial ``cross-model effect.''
Figure~\ref{fig:model-pairs-b} provides a detailed breakdown of acceptance rates for specific model pairings. Each circular marker in the plot represents a unique model pair. 
The radius of each bubble is proportional to the combined parameter size of the models in the pair, with larger bubbles signifying pairings of larger models. 
Figure~\ref{fig:model-pairs-b} shows a clear trend related to model size: larger models generally cluster towards the right side of the plot, signifying higher acceptance rates. This suggests that increasing model scale positively impacts the quality of generated dialogues, as expected. For example, pairs involving large models like LLaMA3-70B and Qwen2.5-72B consistently show high acceptance rates.
Conversely, pairings that include Command-r7B or Gemma3-4B achieve the lowest acceptance rates.

\begin{figure*}
    \centering
    \begin{subfigure}[t]{0.54\textwidth}
        \centering
        \includegraphics[width=\linewidth,height=55mm, keepaspectratio]{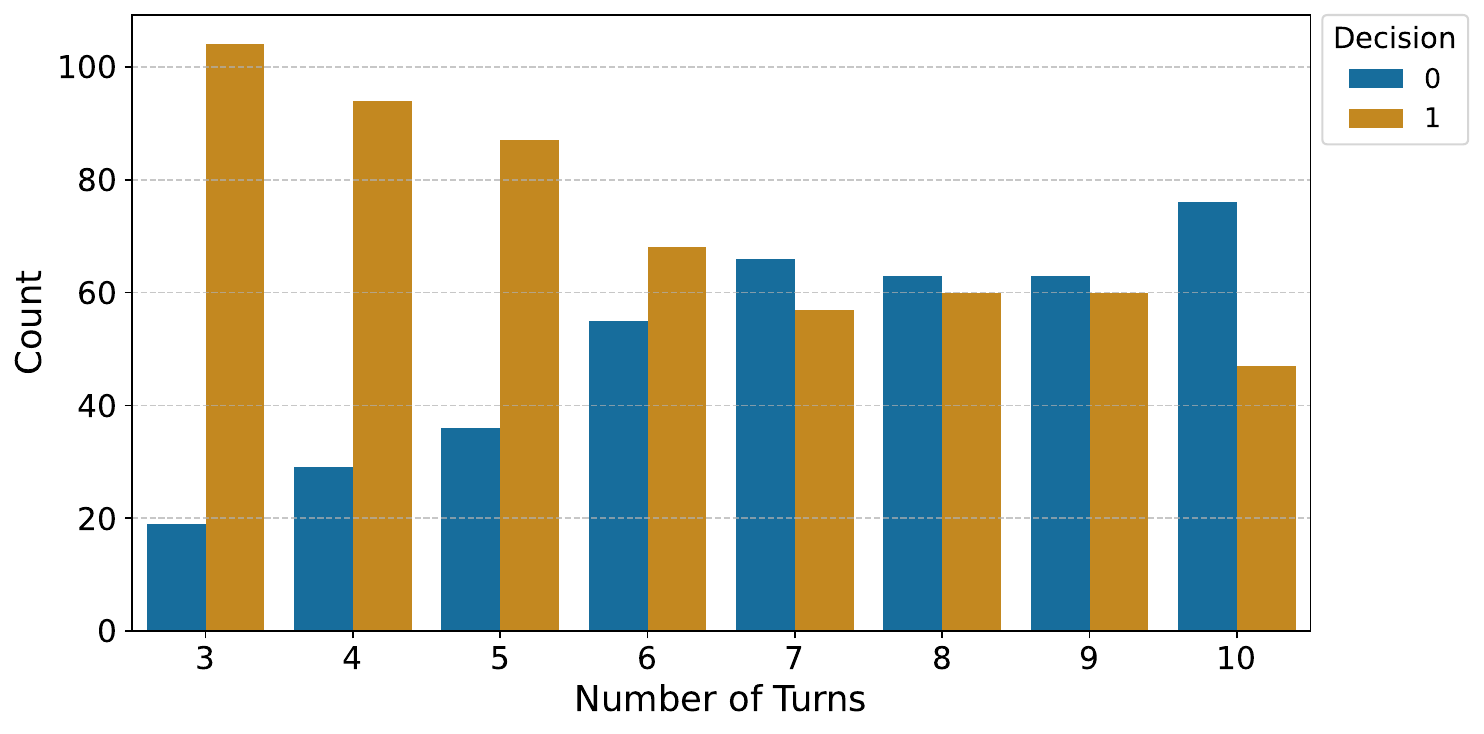}
        \caption{\textit{Number of valid/invalid dialogues\\ per number of turns.}}
        \label{fig:neg-a}
    \end{subfigure}
    \hfill
    \begin{subfigure}[t]{0.41\textwidth}
        \centering
        \includegraphics[width=\linewidth,height=50mm, keepaspectratio]{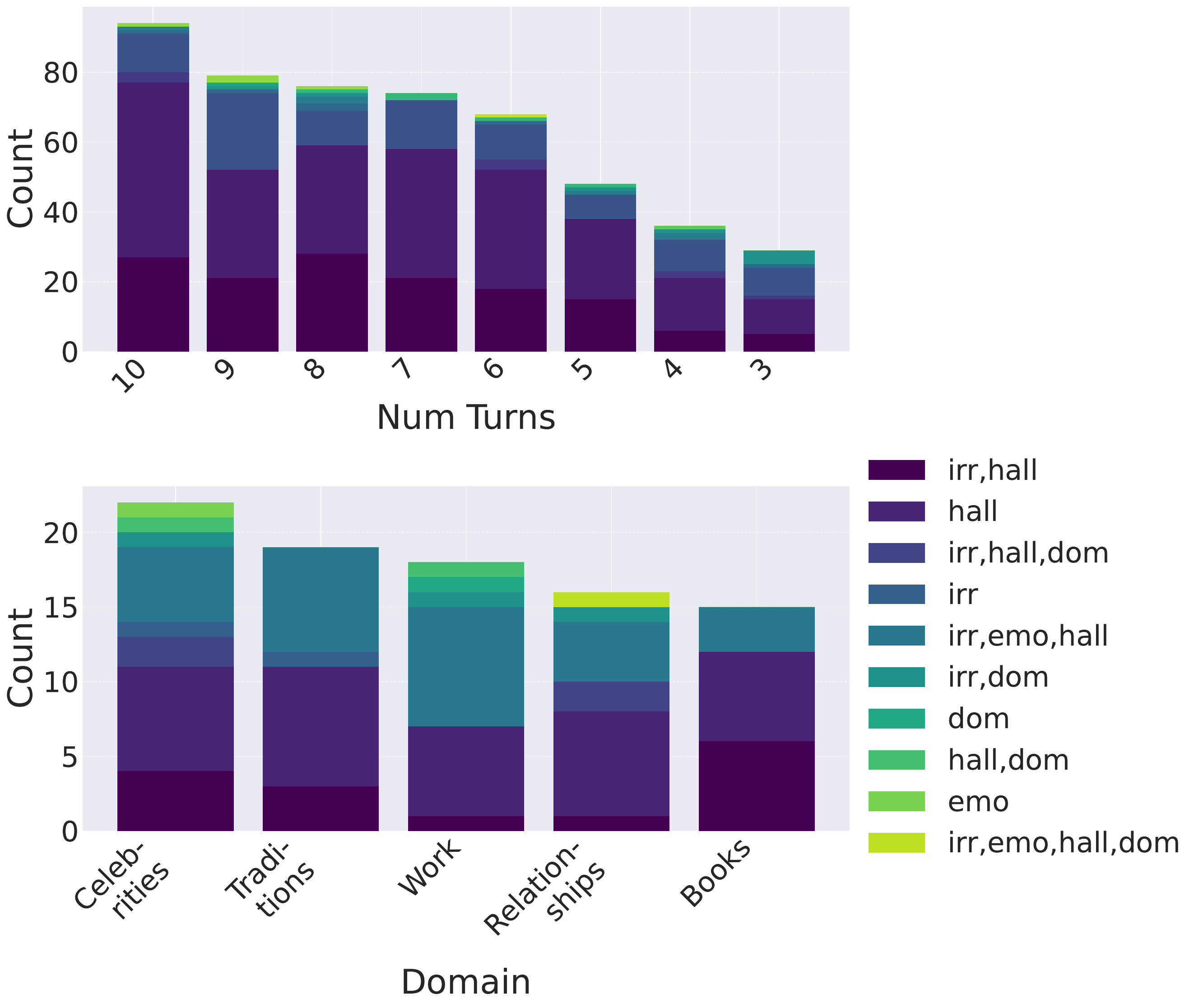}
        \caption{\textit{Invalid dialogues, divided by top-5 domains (top), and number of turns (bottom).}}
        \label{fig:neg-b}
    \end{subfigure}
    \caption{Valid and invalid dialogues (a) and negative reasons (b), as assessed by human annotators.}
    \vspace{-5mm}
    \label{fig:negative-reasons-ab}
\end{figure*}

\textbf{Number of turns distribution}
Figure~\ref{fig:neg-a} shows the distribution of evaluated dialogues based on the number of conversation turns (3-10).
We observe a clear trend: short dialogues ($\leq$5 turns) have significantly more accepted than rejected instances, but this advantage diminishes as conversation length increases.
At 6 turns, the acceptance rate remains higher but with a narrower margin, while for longer dialogues (7-9 turns), the rejection rate becomes comparable to or slightly exceeds the acceptance rate.
At the maximum length of 10 turns, rejected dialogues clearly outnumber accepted ones.
This pattern highlights the increasing challenge of maintaining quality, coherence, and emotional consistency in extended conversations, aligning with recent findings by \cite{laban2025llmslostmultiturnconversation} on LLMs' degrading performance in multi-turn interactions.

\textbf{Invalid dialogues}
To better understand the quality distribution of our dataset, we analyze the subset of dialogues filtered due to negative human judgments, categorized by domain and dialogue length (Figure~\ref{fig:neg-b}). 
The dominant failure modes are hallucination (\textit{hall}) and irrelevance (\textit{irr}), often co-occurring with domain mismatch (\textit{dom}). 
Emotion-related inconsistencies (\textit{emo}), while less common overall, appear more frequently in domains like ``\textit{Relationships}'' and ``\textit{Celebrities}.'' 
We also observe a clear trend with dialogue length: longer conversations (9–10 turns) exhibit a higher rate of failure, underscoring the challenge of maintaining coherence and fidelity over extended interactions. 
These patterns highlight critical areas where current LLMs struggle in multi-turn dialogues. 

A more detailed analysis in the Appendix further confirms that most rejections originate from smaller models, 
particularly when paired with themselves. In contrast, high-capacity models like Qwen2.5-72B and LLaMA3-70B produce substantially fewer invalid dialogues.

\textbf{Concrete vs. abstract domains}

\begin{wrapfigure}{r}{0.5\textwidth} 
    \centering

    \includegraphics[width=0.5\textwidth]{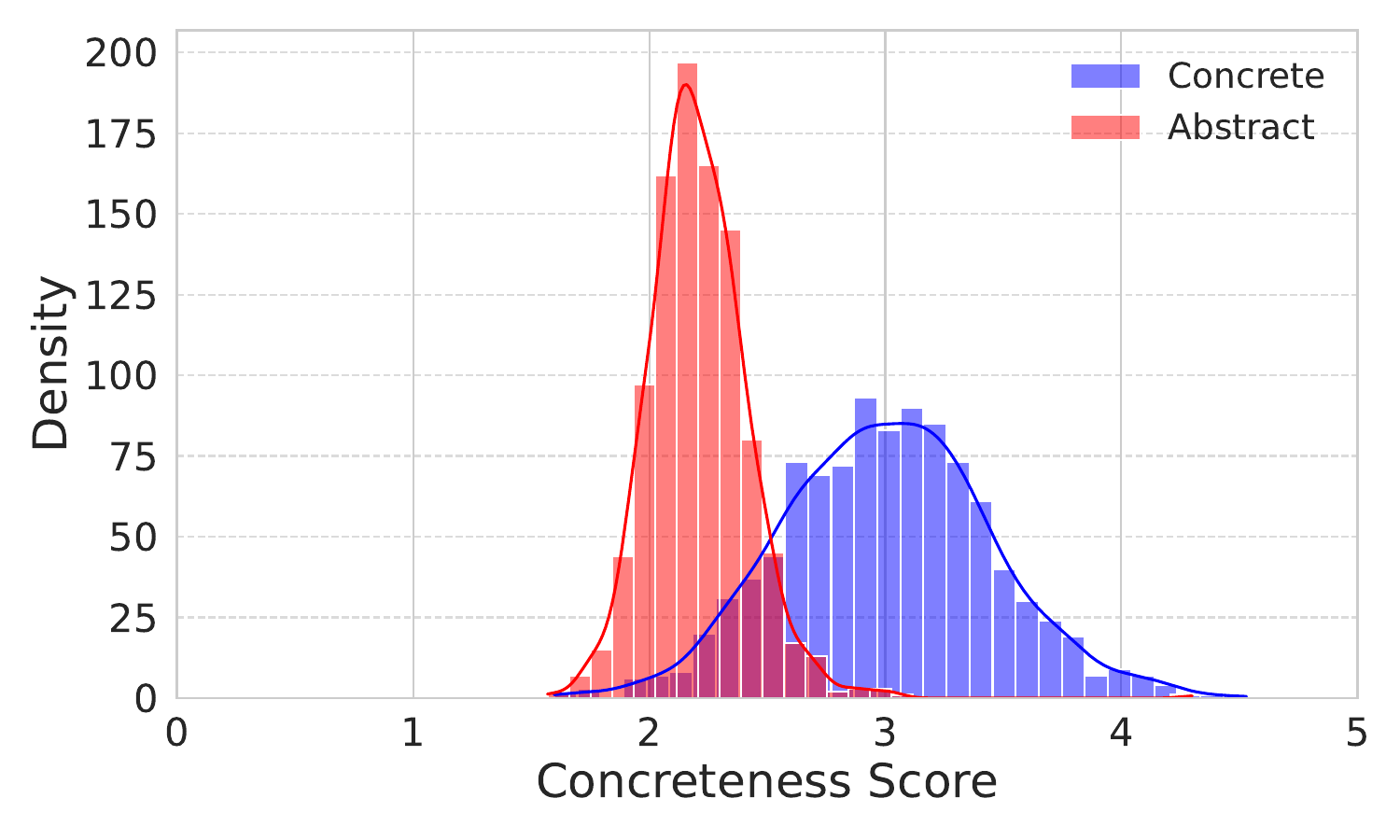}
    \caption{Comparison of concreteness score of concrete vs. abstract accepted dialogues.}
    \label{fig:concretenss}
    \vspace{-3mm}
\end{wrapfigure} 
To quantitatively assess linguistic concreteness in multi-turn dialogues, we leverage the Brysbaert concreteness ratings~\cite{brysbaert2014concreteness}, a widely-used lexicon of 40,000 English words annotated with concreteness scores on a 1–5 scale. 
We use GPT-4o to first parse accepted dialogues across two domain types, Concrete (e.g., ``\textit{Cars}'') and Abstract (e.g., ``\textit{Philosophy}''). 
We randomly sample 1,000 dialogue turns from each domain type and compute the average concreteness score per turn by matching words against~\cite{brysbaert2014concreteness} and averaging the associated concreteness ratings for all matched words.
The resulting distributions, visualized in Figure~\ref{fig:concretenss}, reveal that turns from concrete domains have a higher average concreteness score ($3.01$) than those from abstract domains ($2.21$), highlighting the grounding effect of tangible topics in model-model interaction.

\begin{wraptable}{r}{0.32\textwidth}
\vspace{-4mm}
\caption{Speech emotion recognition performance.}
\vspace{-1mm}
\label{tab:emotion-recognition}
\resizebox{0.32\textwidth}{!}{%
\begin{tabular}{@{}lcc@{}}
\toprule
\textbf{Model} & \textbf{Acc. (\%)}                & \textbf{F1 (\%)}   \\ \midrule
\multicolumn{3}{c}{\textbf{DeepDialogue}}                              \\ \midrule
W2V2-DD  & 88.03                   & 88.10                   \\
WavLM-DD     & 90.20                   & 90.17                   \\
HuBERT-DD    & 93.70                   & 93.64                   \\ \midrule
\multicolumn{3}{c}{\textbf{RAVDESS}}                                   \\ \midrule

HuBERT-DD    & 56.64 & 56.10 \\
HuBERT-LP & 65.28                   & /                       \\ \bottomrule
\end{tabular}%
}
\vspace{-4mm}
\end{wraptable}

\subsection{Speech Emotion Recognition}
While DeepDialogue is mainly designed to support research in conversational AI, we also investigate the extent to which its generated dialogues preserve emotional consistency. 
To this end, we construct a balanced subset of the \texttt{XTTS-v2} dataset comprising 1,000 dialogue turns per emotion category and use it to train three self-supervised learning (SSL) models, Wav2Vec2~\cite{wav2vec2}, WavLM~\cite{wavlm}, and HuBERT~\cite{hubert}, for speech emotion recognition. All models demonstrate strong performance (around 90\%) in both accuracy and macro F1-score when evaluated on a held-out test set from DeepDialogue.
To assess the generalizability of the learned representations, we evaluate the best-performing model in a zero-shot setting on the RAVDESS dataset~\cite{ravdess}, which shares the same emotion label distribution. As expected, we observe a performance drop due to domain shift; however, results remain strong. Our zero-shot HuBERT-DD achieves an accuracy of 56.6\%, closely approaching the performance of a linear-probing baseline (HuBERT-LP, 65.3\%) trained directly on RAVDESS~\cite{arch}. This suggests that the emotional content in DeepDialogue is not only consistent but also transferable, supporting its utility as a high-quality resource for emotion-aware dialogues. 

\section{Conclusions}
~\label{section:conclusion}
We presented DeepDialogue, a large-scale multimodal dataset containing 40,150 high-quality multi-turn dialogues across 41 domains with 20 distinct emotions and coherent emotional progressions.
Key findings include the degradation of smaller models' performance beyond 6 turns, superior dialogue quality in concrete domains compared to abstract ones, and the beneficial impact of cross-model interactions on coherence.
Our dual approach to speech synthesis provides researchers with complementary spoken utterances that capture both explicit emotion-conditioned and implicit linguistically-derived emotional expressions in conversational speech.
DeepDialogue bridges critical gaps in dialogue research by providing a comprehensive resource that integrates emotional consistency, domain diversity, conversation depth, and multimodality.
The speech component enables novel research at the intersection of text-based dialogue systems and speech-based conversational AI, addressing the need for emotionally expressive multimodal dialogue data.
Future work will focus on leveraging this resource to fine-tune dialogue models to enhance emotional intelligence and conversational coherence, develop more sophisticated emotion transition models, and extend the framework to additional languages and cultural contexts.

\textbf{Limitations}~\label{section:limitations}
DeepDialogue addresses several gaps in existing literature, though inherent limitations remain challenging to overcome with current technologies.
The synthetic nature of our dataset, despite rigorous quality filtering, may differ from authentic human interactions, especially in longer conversations where maintaining coherence becomes difficult even for state-of-the-art models.
Our emotion taxonomy, though grounded in psychological literature, represents a practical simplification of human emotional expression, which in reality exists on continuous spectra rather than as discrete categories.
Regarding speech synthesis, current technology fundamentally cannot replicate all the subtle vocal characteristics of human emotional expression, such as spontaneous voice breaks, natural hesitations, and the physiological effects of genuine emotions on vocal production.
The dataset inherently reflects the distribution patterns of the underlying language models, which, despite our diverse domain selection, may not equally represent all cultural contexts or conversational styles.
Finally, while our human-LLM hybrid evaluation achieves strong agreement metrics, some aspects of conversational naturalness remain subjective and difficult to quantify systematically.

\noindent \textbf{Ethical Considerations}
\label{section:ethical_considerations} 
DeepDialogue raises ethical considerations regarding representation bias, as the synthetic dialogues inevitably reflect biases present in the underlying language models' training data.
Emotional speech synthesis also presents concerns regarding potential misuse for creating misleading content designed to elicit specific emotional responses.
Our dataset consists entirely of synthetic dialogues rather than recordings of real human conversations, which eliminates direct privacy concerns associated with human data collection.
We acknowledge these ethical challenges while providing this resource to advance research in emotionally intelligent conversational AI.

\bibliographystyle{plain}
\bibliography{neurips_2025}

\appendix

\section{Dataset Creation}\label{appendix:creation}
This section outlines the full pipeline for constructing the \dd dataset, detailing how emotionally rich, domain-grounded dialogues are generated, evaluated, and converted into speech.
Section \S\ref{appendix:schema} describes the four-stage pipeline, from LLM-based generation and human evaluation to automated filtering and multimodal synthesis.
Section \S\ref{appendix:domain} introduces the 41 domains used, categorized by concreteness and balanced across dialogues to support broad contextual diversity.
Section \S\ref{appendix:emotion} explains the modeling of 20 emotion categories using transition graphs and domain-specific mappings to ensure emotional plausibility and coherence.
Section \S\ref{appendix:prompt} details prompt templates used to guide LLM behavior across dialogue turns, ensuring emotional grounding and natural conversational style.
Finally, Section \S\ref{appendix:TTS} covers the text-to-speech conversion using \texttt{XTTS-v2} with emotion conditioning and \texttt{Orpheus} with implicit expression, along with preprocessing steps for clean synthesis input.

\subsection{Generation Framework}
\label{appendix:schema}

Figure~\ref{fig:dd-framework} provides a visual summary of the \dd creation pipeline.
Unlike previous datasets, our framework integrates model-based dialogue generation with multi-stage quality filtering and multimodal extension.
The pipeline consists of four interconnected components.

The foundation of our approach pairs 9 different language models (4B-72B parameters) to generate conversations across 41 domains and 20 emotion categories.
Each dialogue covers between 3-10 turns, with models instructed to maintain domain relevance while expressing specific emotions.
This process yielded 65,600 initial dialogues with diverse conversational dynamics and emotional progressions.

From the generated pool, with stratified sampling, we randomly selected 984 dialogues for human evaluation (123 per turn type, 24 per domain).
Three independent annotators assessed each conversation for coherence, emotional consistency, presence of hallucination, and domain adherence.
These expert judgments achieved substantial agreement (Fleiss' $\kappa = 0.80$), establishing a robust foundation for scaling our quality standards.

Building on human annotations, we selected an ensemble of open-source LLMs to extend quality assessment to all 65,600 dialogues.
The ensemble closely mirrored human judgment patterns (Cohen's $\kappa = 0.79$), enabling reliable automated filtering.
After applying consistent quality thresholds and safety checks, the filtered dataset contained 40,150 high-quality dialogues, representing the 61\% of the initial collection.

The filtered text dialogues were finally synthesized into spoken conversations through two complementary speech synthesis approaches.
\texttt{XTTS-v2}~\cite{xtts} leverages emotional conditioning using reference samples from the RAVDESS~\cite{ravdess} dataset, explicitly mapping our 20 emotions to appropriate vocal expressions.
On the other hand, \texttt{Orpheus}~\cite{orpheus} generates highly natural speech relying solely on linguistic cues for implicit emotional expression.
Both methods produced consistent speaker identities throughout each dialogue, resulting in over 480 hours of audio per variant.

This integrated framework creates a continuous pipeline from generation to multimodal extension, enabling systematic investigation of dialogue quality factors while bridging text-based and speech-based conversational AI research.

\begin{table}[]
      \caption{\dd's 41 domains, detailed descriptions, concrete vs. abstract categories (as assessed by GPT-4o), and overall final count.} 
      \label{table-all-domains}
      \centering
        \scalebox{0.77}{%
        \begin{tabular}{lllc}
        \toprule
        \textbf{Domain}              & \textbf{Description}                                                             & \textbf{Category}    & \textbf{Count} \\
        \midrule            
        \texttt{Art}                 & Discussing paintings, sculptures, or artistic movements                          & Abstract             & 1041               \\
        \texttt{Books}               & Discussing literature, authors, or reading preferences                           & Concrete             & 999               \\
        \texttt{Cars}                & Discussing car models, driving experiences, or automotive technology             & Concrete             & 1035               \\
        \texttt{Celebrities}         & Discussing famous personalities, movies, or music                                & Concrete             & 852               \\
        \texttt{Coding}              & Discussing programming languages, software development, or coding challenges     & Concrete             & 970 \\
        \texttt{Cooking}             & Discussing recipes, cooking techniques, or kitchen gadgets                       & Concrete             & 928               \\       
        \texttt{Education}           & Discussing learning methods, subjects, or educational systems                    & Abstract             & 1063               \\
        \texttt{Events}              & Discussing social gatherings, festivals, or special occasions                    & Concrete             & 1026               \\
        \texttt{Fashion}             & Discussing clothing styles, fashion trends, or personal aesthetics               & Concrete             & 1033 \\
        \texttt{Fitness}             & Discussing workout routines, gym equipment, or fitness goals                     & Concrete             & 898               \\
        \texttt{Finance}             & Discussing investments, budgeting, or financial planning                         & Abstract             & 769               \\
        \texttt{Food}                & Discussing recipes, restaurants, or culinary preferences                         & Concrete             & 992               \\
        \texttt{Gaming}              & Discussing video games, consoles, or gaming strategies                           & Concrete             & 927               \\
        \texttt{Gardening}           & Discussing plant care, gardening tools, or landscaping ideas                     & Concrete             & 1006               \\
        \texttt{Health}              & Discussing fitness routines, diets, or medical topics                            & Concrete             & 893               \\
        \texttt{History}             & Discussing historical events, figures, or time periods                           & Concrete             & 1039               \\
        \texttt{Hobbies}             & Discussing interests, hobbies, or leisure activities                             & Concrete             & 960               \\
        \texttt{Holidays}            & Discussing festive celebrations, holiday destinations, or seasonal activities    & Concrete             & 1058 \\
        \texttt{Home}                & Discussing interior design, home decor, or DIY projects                          & Concrete             & 1015               \\
        \texttt{Languages}           & Discussing language learning, linguistic diversity, or translation               & Abstract             & 1087 \\
        \texttt{Makeup}              & Discussing makeup techniques, beauty products, or cosmetic brands                & Concrete             & 990               \\
        \texttt{Movies}              & Discussing favorite films, directors, or recent releases                         & Concrete             & 866               \\
        \texttt{Music}               & Discussing bands, songs, or concert experiences                                  & Concrete             & 988               \\
        \texttt{Nature}              & Discussing outdoor activities, environmental issues, or natural wonders          & Concrete             & 1083 \\
        \texttt{News}                & Discussing actualities, news sources, or current events                          & Abstract             & 1028               \\
        \texttt{Pets}                & Discussing pet care, animal behavior, or favorite animals                        & Concrete             & 1009               \\
        \texttt{Philosophy}          & Discussing existential questions or philosophical concepts                       & Abstract             & 1071               \\
        \texttt{Photography}         & Discussing camera gear, photography tips, or photo editing                       & Concrete             & 982               \\
        \texttt{Podcasts}            & Discussing favorite podcasts, episodes, or podcast hosts                         & Concrete             & 965               \\
        \texttt{Politics}            & Discussing current events, political figures, or ideologies                      & Abstract             & 959               \\
        \texttt{Relationships}       & Discussing dating experiences, friendships, or family dynamics                   & Abstract             & 851               \\
        \texttt{Science}             & Discussing scientific discoveries or theories                                    & Abstract             & 1065               \\
        \texttt{Shopping}            & Discussing shopping habits, fashion trends, or online purchases                  & Concrete             & 995               \\
        \texttt{Social Media}        & Discussing online platforms, influencers, or viral trends                        & Concrete             & 1095               \\
        \texttt{Spirituality}        & Discussing meditation, mindfulness, or spiritual practices                       & Abstract             & 1025               \\
        \texttt{Sports}              & Discussing teams, athletes, or recent sports events                              & Concrete             & 1043               \\
        \texttt{Technology}          & Discussing new gadgets, software, or tech trends                                 & Concrete             & 923               \\
        \texttt{Traditions}          & Discussing cultural customs, holidays, or family traditions                      & Abstract             & 956               \\
        \texttt{Travel}              & Discussing favorite destinations, travel experiences, or planning trips          & Concrete             & 984 \\
        \texttt{Weather}             & Discussing climate patterns, weather forecasts, or natural disasters             & Concrete             & 991               \\
        \texttt{Work}                & Discussing job responsibilities, career goals, or workplace                      & Concrete             & 690               \\
        \bottomrule
        \end{tabular}
    }
        
\end{table}
\subsection{Domains}\label{appendix:domain}
To ensure a wide range of emotionally expressive conversations, \dd spans 41 distinct domains, each associated with everyday topics that people commonly discuss in informal settings. These domains were carefully curated to reflect a broad spectrum of interests and contexts, ranging from tangible, everyday subjects like \texttt{Food}, \texttt{Cars}, or \texttt{Pets}, to more abstract or conceptual areas such as \texttt{Philosophy}, \texttt{Spirituality}, or \texttt{Education}.

\begin{figure}
    \centering
    \includegraphics[width=0.85\linewidth]{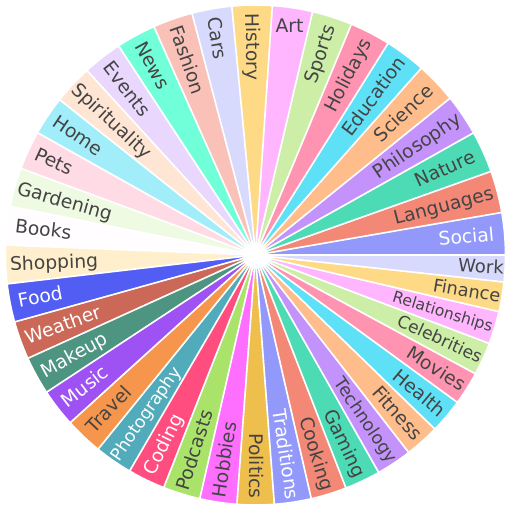}
    \caption{Domain distribution in \ddnospace.}
    \label{fig:sunburst-domains}
\end{figure}
Table~\ref{table-all-domains} provides an overview of all domains included in the final released dataset, along with concise descriptions, a high-level classification into either \textit{concrete} or \textit{abstract} (as assessed by GPT-4o), and the number of dialogue instances per domain. 

Domain selection during generation followed a stratified sampling strategy to ensure a perfectly balanced representation across all domains, which eliminates domain-based sampling bias and supports robust cross-domain comparisons. Figure~\ref{fig:sunburst-domains} shows the final distribution of the domains in \dd. 

\begin{table}[]
      \caption{\dd's emotion and relative chains.} 
      \label{table-all-emotions}
      \centering
        \scalebox{0.95}{%
        \begin{tabular}{ll}
        \toprule
        \textbf{Emotion}           & \textbf{Chain}                                     \\
        \midrule
        \texttt{Amused}       & [Happy, Excited, Relaxed, Surprised, Curious]                    \\  
        \texttt{Angry}        & [Frustrated, Disappointed, Surprised, Anxious, Worried]          \\
        \texttt{Bored}        & [Curious, Frustrated, Relaxed, Surprised]                        \\
        \texttt{Anxious}      & [Worried, Confused, Frustrated, Hopeful, Surprised]              \\
        \texttt{Confused}     & [Curious, Worried, Frustrated, Surprised, Anxious]               \\
        \texttt{Curious}      & [Surprised, Excited, Confused, Amused, Enthusiastic]             \\
        \texttt{Disappointed} & [Sad, Frustrated, Anxious, Hopeful, Angry]                       \\
        \texttt{Embarrassed}  & [Anxious, Worried, Amused, Frustrated, Surprised]                \\
        \texttt{Enthusiastic} & [Excited, Happy, Curious, Hopeful, Proud]                        \\
        \texttt{Excited}      & [Happy, Surprised, Enthusiastic, Curious, Proud, Amused]            \\
        \texttt{Frustrated}   & [Angry, Disappointed, Anxious, Worried, Surprised]               \\
        \texttt{Grateful}     & [Happy, Relaxed, Hopeful, Proud, Nostalgic]                      \\
        \texttt{Happy}        & [Excited, Proud, Amused, Grateful, Relaxed, Enthusiastic, Curious]  \\
        \texttt{Hopeful}      & [Excited, Happy, Curious, Grateful, Anxious, Nostalgic]             \\
        \texttt{Nostalgic}    & [Happy, Sad, Relaxed, Grateful, Hopeful]                         \\
        \texttt{Proud}        & [Happy, Excited, Grateful, Enthusiastic, Relaxed]                \\
        \texttt{Relaxed}      & [Happy, Amused, Nostalgic, Grateful, Hopeful]                    \\
        \texttt{Sad}          & [Disappointed, Worried, Nostalgic, Frustrated, Anxious, Hopeful]    \\    
        \texttt{Surprised}    & [Excited, Confused, Curious, Amused, Worried, Happy]                \\  
        \texttt{Worried}      & [Anxious, Confused, Sad, Frustrated, Hopeful]                    \\        
        \bottomrule
        \end{tabular}}
\end{table}
\begin{figure}
    \centering
    \includegraphics[width=0.90\linewidth]{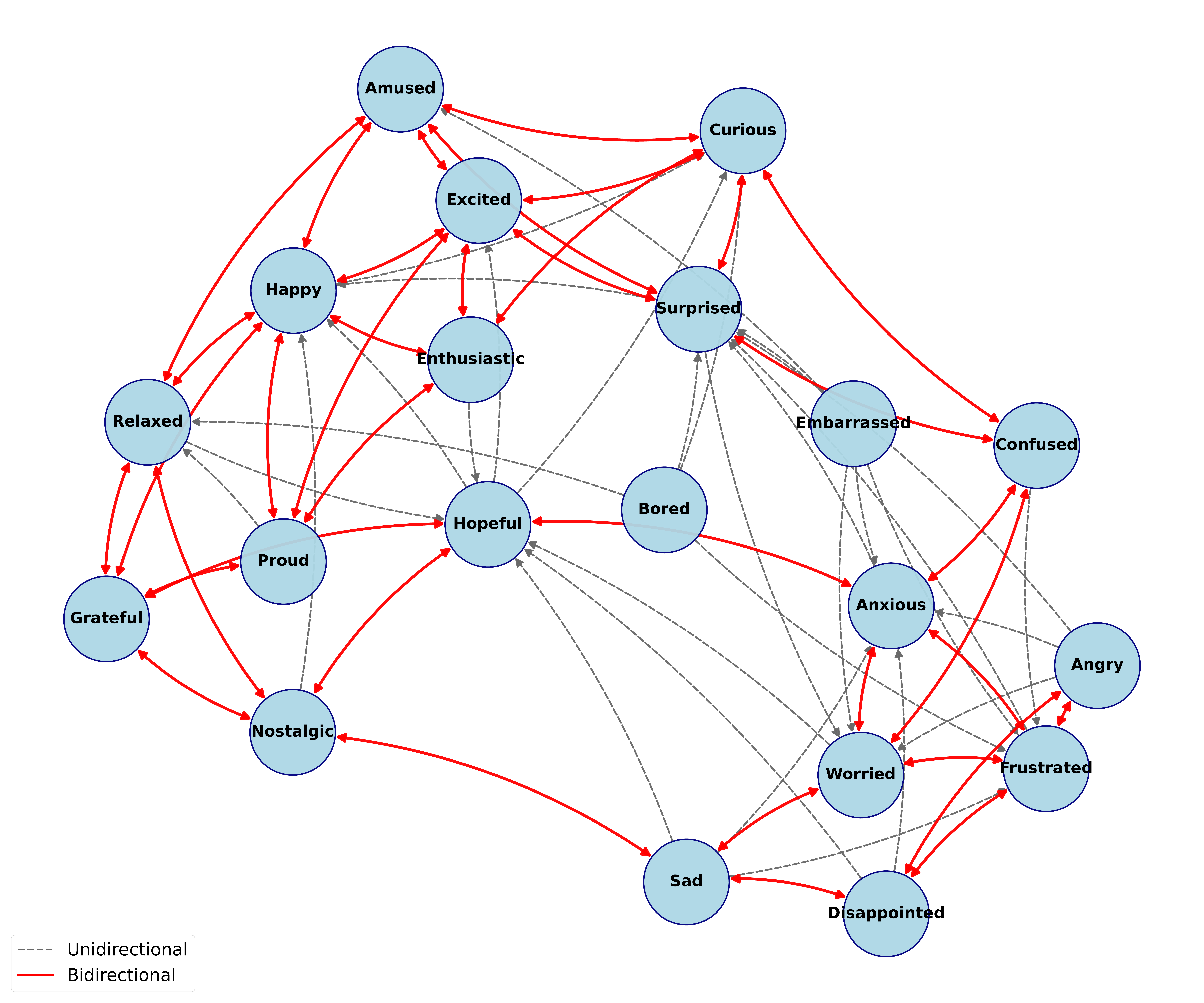}
    \caption{Emotion transition graph in \dd.}
    \label{fig:emotion-transition-graph}
\end{figure}
\subsection{Emotional Progress}\label{appendix:emotion}

\textbf{Emotion Chains: Modeling Natural Transitions.} The emotion chains in Table~\ref{table-all-emotions} (and graphically shown in Figure~\ref{fig:emotion-transition-graph}) establish a \textit{directed graph} of plausible emotional transitions. Each emotion is associated with a set of semantically or psychologically plausible next emotions. For instance, \texttt{Frustrated} is associated with \textit{Angry, Disappointed, Anxious, Worried, Surprised}, \texttt{Curious} with \textit{Surprised, Excited, Confused, Amused, Enthusiastic}, and \texttt{Sad} with \textit{Disappointed, Worried, Nostalgic, Frustrated, Anxious, Hopeful}.  
These reflect emotion dynamics grounded in psychological theory, particularly from Russell’s circumplex model of affect~\cite{russell1980circumplex} and Plutchik’s emotion wheel~\cite{plutchik2001nature}. The chains support: (i) gradual transitions (e.g., \textit{Confused} $\rightarrow$ \textit{Curious} $\rightarrow$ \textit{Surprised} $\rightarrow$ \textit{Excited}), (ii) valence-constrained paths (e.g., rarely jumping from \textit{Sad} $\rightarrow$ \textit{Happy} directly), and (iii) recovery patterns (e.g., \textit{Sad} $\rightarrow$ \textit{Hopeful} or \textit{Anxious} $\rightarrow$ \textit{Hopeful}). 
By constraining dialogue flow to these chains, we ensure emotional coherence and realism.

\begin{table}[]
      \caption{\dd's mapping emotion-domains.} 
      \label{table-emotion-domain-mapping}
      \centering
        \scalebox{0.80}{%
        \begin{tabular}{llc}
        \toprule
        \textbf{Domain}   & \textbf{Emotion mapping}                                                                         \\
        \midrule
        \texttt{Art}             & [Happy, Excited, Curious, Proud, Amused, Enthusiastic, Hopeful]                \\
        \texttt{Books}           & [Curious, Excited, Nostalgic, Amused, Relaxed, Enthusiastic, Surprised, Proud, Sad]                    \\
        \texttt{Cars}            & [Excited, Happy, Curious, Proud, Amused, Enthusiastic, Hopeful]                \\
        \texttt{Celebrities}     & [Curious, Excited, Amused, Grateful, Sad, Enthusiastic, Frustrated, Surprised, Angry]                 \\
        \texttt{Coding}          & [Excited, Happy, Curious, Proud, Amused, Enthusiastic, Hopeful]                \\
        \texttt{Cooking}         & [Happy, Excited, Amused, Grateful, Relaxed, Enthusiastic, Curious]             \\
        \texttt{Education}       & [Curious, Excited, Hopeful, Proud, Enthusiastic, Amused, Relaxed, Angry]     \\
        \texttt{Events}          & [Happy, Excited, Amused, Grateful, Proud, Enthusiastic, Curious, Sad]        \\
        \texttt{Fashion}         & [Excited, Happy, Curious, Proud, Amused, Enthusiastic, Hopeful]                \\
        \texttt{Finance}         & [Worried, Anxious, Hopeful, Curious, Confused, Frustrated, Surprised, Angry] \\
        \texttt{Fitness}         & [Happy, Excited, Amused, Grateful, Relaxed, Enthusiastic, Curious, Surprised, Frustrated, Angry]     \\
        \texttt{Food}            & [Happy, Excited, Curious, Amused, Grateful, Enthusiastic, Nostalgic]           \\
        \texttt{Gaming}          & [Excited, Happy, Curious, Proud, Amused, Enthusiastic, Hopeful]                \\
        \texttt{Gardening}       & [Happy, Excited, Amused, Grateful, Relaxed, Enthusiastic, Curious]             \\
        \texttt{Health}          & [Hopeful, Happy, Grateful, Relaxed, Excited, Proud, Curious, Sad]            \\
        \texttt{History}         & [Nostalgic, Curious, Happy, Proud, Amused, Relaxed, Grateful]                  \\
        \texttt{Hobbies}         & [Happy, Excited, Amused, Grateful, Relaxed, Enthusiastic, Curious]             \\
        \texttt{Holidays}        & [Happy, Excited, Amused, Grateful, Relaxed, Enthusiastic, Curious, Sad]      \\
        \texttt{Home}            & [Happy, Excited, Amused, Grateful, Relaxed, Enthusiastic, Curious]             \\
        \texttt{Languages}       & [Curious, Excited, Hopeful, Proud, Enthusiastic, Amused, Relaxed]              \\
        \texttt{Makeup}          & [Happy, Excited, Amused, Grateful, Relaxed, Enthusiastic, Curious]             \\
        \texttt{Movies}          & [Excited, Amused, Surprised, Disappointed, Enthusiastic, Nostalgic, Curious, Sad]                        \\
        \texttt{Music}           & [Happy, Excited, Nostalgic, Relaxed, Amused, Enthusiastic, Hopeful, Sad]     \\
        \texttt{Nature}          & [Happy, Excited, Amused, Grateful, Relaxed, Enthusiastic, Curious]            \\
        \texttt{News}            & [Happy, Excited, Amused, Grateful, Relaxed, Enthusiastic, Curious, Surprised, Sad, Worried, Angry] \\
        \texttt{Pets}            & [Happy, Excited, Amused, Grateful, Relaxed, Enthusiastic, Curious]             \\
        \texttt{Philosophy}      & [Curious, Confused, Hopeful, Surprised, Excited, Amused, Relaxed]              \\
        \texttt{Photography}     & [Happy, Excited, Amused, Grateful, Relaxed, Enthusiastic, Curious]             \\
        \texttt{Podcasts}        & [Happy, Excited, Amused, Grateful, Relaxed, Enthusiastic, Curious, Sad]     \\
        \texttt{Politics}        & [Angry, Frustrated, Hopeful, Surprised, Curious, Confused, Worried]            \\
        \texttt{Relationships}   & [Happy, Excited, Amused, Grateful, Relaxed, Enthusiastic, Curious, Sad]      \\
        \texttt{Scienc}e         & [Curious, Excited, Hopeful, Proud, Enthusiastic, Amused, Relaxed]              \\
        \texttt{Shopping}        & [Happy, Excited, Amused, Grateful, Relaxed, Enthusiastic, Curious]             \\
        \texttt{Social media}    & [Happy, Excited, Amused, Grateful, Relaxed, Enthusiastic, Curious, Sad]     \\
        \texttt{Spirituality}    & [Happy, Excited, Amused, Grateful, Relaxed, Enthusiastic, Curious, Sad]     \\
        \texttt{Sports}          & [Excited, Happy, Proud, Enthusiastic, Hopeful, Amused, Relaxed]                \\
        \texttt{Technology}      & [Excited, Curious, Surprised, Confused, Enthusiastic, Proud, Frustrated]       \\
        \texttt{Traditions}      & [Happy, Excited, Amused, Grateful, Relaxed, Enthusiastic, Curious]             \\
        \texttt{Travel}          & [Excited, Happy, Curious, Nostalgic, Enthusiastic, Hopeful, Relaxed, Surprised, Amused, Sad]         \\    
        \texttt{Weather}         & [Happy, Excited, Curious, Proud, Amused, Enthusiastic, Hopeful]                \\
        \texttt{Work}            & [Frustrated, Sad, Anxious, Hopeful, Curious, Confused, Excited, Surprised]   \\ 
        \bottomrule
        \end{tabular}}
\end{table}
\begin{figure}
    \centering
    \includegraphics[width=0.85\linewidth]{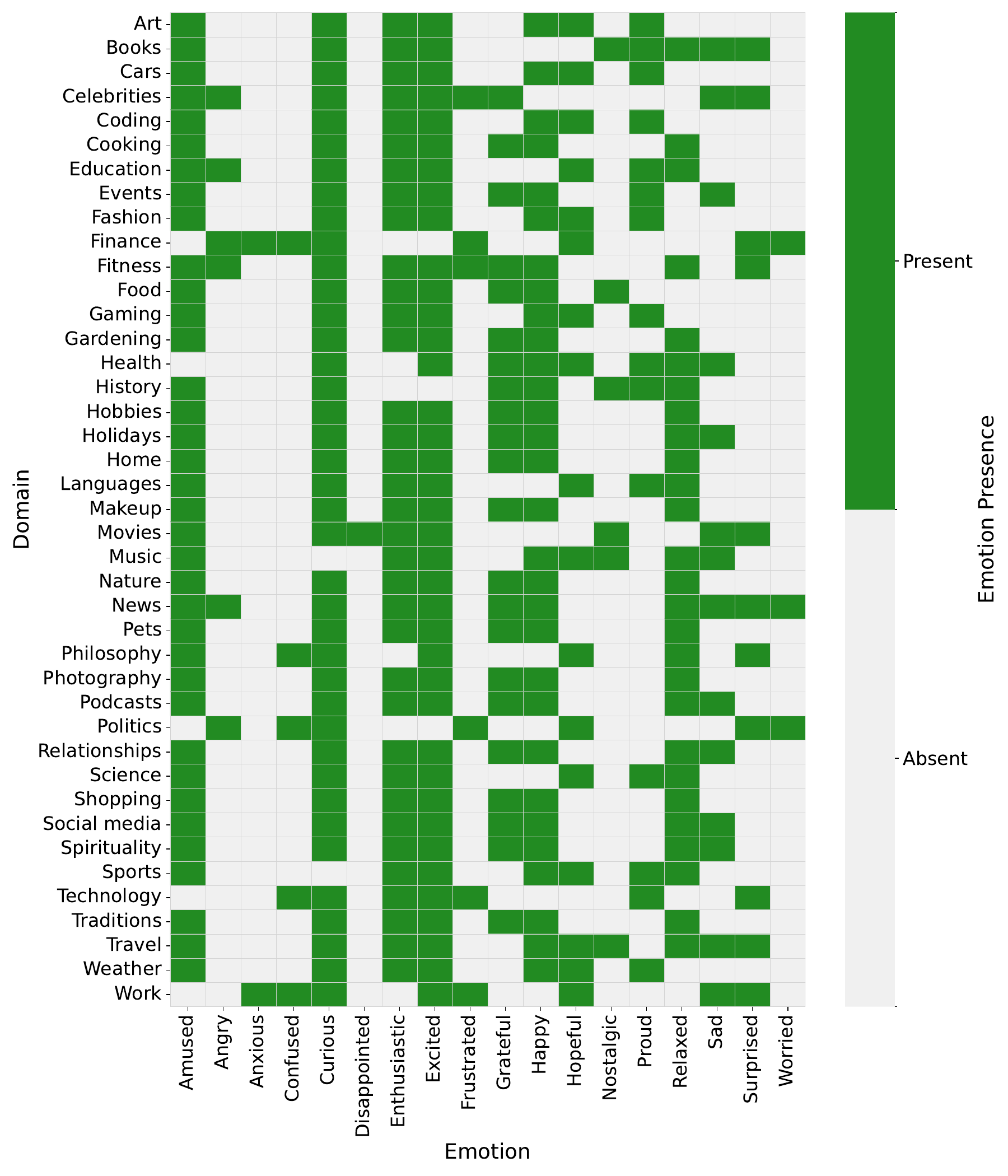}
    \caption{Heatmap of emotion-domain mapping in \ddnospace. Presence of an emotion for a specific domain highlighted in forest green, absence in light gray.}
    \label{fig:domain-emotion-heatmap}
    \vspace{-9mm}
\end{figure}
\textbf{Emotion-Domain Mapping: Grounding Emotions in Context.} Table~\ref{table-emotion-domain-mapping} maps domain-specific emotional expressions, identifying which emotions are plausible or typical in each context. 
This setup allows us to model:
\begin{itemize}
    \item \textit{Semantic grounding}: Domains like \texttt{Finance} include emotions as \textit{Anxious, Worried, Confused}, which are relevant to uncertainty and risk.
    \item \textit{Affective diversity:} Entertainment domains (\texttt{Movies}, \texttt{Music}) span both positive (\textit{Amused, Excited}) and negative (\textit{Sad, Disappointed}) emotions.
    \item \textit{Emotion density and diversity:} Some domains support a broader range of emotional expressions than others. For instance, domains like \texttt{Movies}, \texttt{News}, and \texttt{Travel} feature both epistemic (e.g., \textit{Curious}, \textit{Surprised}) and social emotions (e.g., \textit{Grateful}, \textit{Sad}). In contrast, more specialized domains like \texttt{Finance} or \texttt{Technology} display a narrower emotional palette.
\end{itemize}
This mapping serves as a domain-specific emotional prior, filtering which emotions can be initially sampled or transitioned into. It also prevents emotionally implausible pairings. Figure~\ref{fig:domain-emotion-heatmap} visually depicts such mapping.

\textbf{Emotional Progression in Dialogue Generation.} 
The dialogue generation pipeline incorporates emotion in a multi-stage, probabilistic, and content-aware fashion. 
At the initial state, an emotion is sampled from the set relevant to the selected domain.
At each turn, the next emotion is chosen using the directed transition graph, modulated by the previous emotional state, the current domain, the sentiment cues in the previous message, and some variability. Transitions are not deterministic; instead, weights introduce stochasticity while preventing abrupt, implausible emotional shifts.

Short dialogues favor emotionally tight arcs (e.g., \textit{Curious} $\rightarrow$ \textit{Surprised} $\rightarrow$ \textit{Excited}).
Longer dialogues may instead include negative-positive shifts or compound trajectories (e.g., \textit{Frustrated} $\rightarrow$ \textit{Disappointed} $\rightarrow$ \textit{Hopeful} $\rightarrow$ \textit{Grateful}).
This enables the simulation of emotional narratives and human-like expressive variation across dialogue lengths and model types.

\subsection{Conversational Prompts for LLMs}
\label{appendix:prompt}

\begin{figure}[t]
    \centering
    \footnotesize
    \begin{minipage}{\columnwidth}
    
    \textbf{Model prompt, initial turn.}
        \begin{tcolorbox}[promptbox]
            \scriptsize
            \textcolor{systemgray}{[System]} 
            \newline
            \textcolor{promptblue}{You are an AI assistant for text generation with human sentiments.
            \newline
            You are given the information about the domain and the initial emotion.
            \newline
            You have to provide an answer as you would talk to a close friend, showing authentic emotion.}
            \newline
            \newline
            \textcolor{inputred}{NOTE:}
            \newline
            \textcolor{promptblue}{Do NOT include any special tokens, prefixes, or suffixes in your response.
            \newline
            Do NOT include the prompt in your response.
            \newline
            Strictly follow the instructions.}
            \newline
            \newline
            \textcolor{inputred}{EXAMPLE:}
            \newline
            \textcolor{promptblue}{For example, if the domain is CARS, an answer would be: ``Have you heard about the new Tesla model?''.} 
            \newline
            \newline
            \textcolor{inputred}{INPUT:}
            \newline
            \textcolor{promptblue}{You are having a brief emotional conversation about} \textcolor{inputgreen}{ \{DOMAIN\_DESCRIPTION\}}\textcolor{promptblue}{.
            \newline
            Your emotional state:} \textcolor{inputgreen}{\{INITIAL\_EMOTION\}}\textcolor{promptblue}{.}
            \newline
            \newline
            \textcolor{inputred}{OUTPUT (MAX 25 Words):}
        \end{tcolorbox}
        
        \textbf{Model prompt, from second turn on.}
        \begin{tcolorbox}[promptbox]
            \scriptsize
            \textcolor{systemgray}{[System]} 
            \newline
            \textcolor{promptblue}{You are an AI assistant for text generation with human sentiments.
            \newline
            You are given the information about the domain, the conversation so far, and the next emotion.
            \newline
            You have to provide an answer as you would talk to a close friend, showing authentic emotion.}
            \newline
            \newline
            \textcolor{inputred}{NOTE:}
            \newline
            \textcolor{promptblue}{Do NOT include any special tokens, prefixes, or suffixes in your response.
            \newline
            Do NOT include the prompt in your response.
            \newline
            Strictly follow the instructions.}
            \newline
            \newline
            \textcolor{inputred}{EXAMPLE:}
            \newline
            \textcolor{promptblue}{For example, if the domain is CARS and the context is ``Have you heard about the new Tesla model?'', an answer would be: ``Oh yes, I saw the announcement yesterday, it seems really impressive!''.} 
            \newline
            \newline
            \textcolor{inputred}{INPUT:}
            \newline
            \textcolor{promptblue}{You are having a brief emotional conversation about} \textcolor{inputgreen}{ \{DOMAIN\_DESCRIPTION\}}\textcolor{promptblue}{.
            \newline
            Your emotional state:} \textcolor{inputgreen}{\{NEXT\_EMOTION\}}\textcolor{promptblue}{.
            \newline
            Respond naturally to the conversation so far:} \textcolor{inputgreen}{\{CONTEXT\}}
            \newline
            \newline
            \textcolor{inputred}{OUTPUT (MAX 25 Words):}
        \end{tcolorbox}
        
    \end{minipage}
    \caption{Example of model prompts for dialogue generation. The placeholders in curly brackets for \texttt{DOMAIN}, \texttt{EMOTION}, and \texttt{CONTEXT} are dynamically filled based on the sampling strategy.}
    \label{fig:prompt-templates}
    \vspace{-3mm}
\end{figure}

To generate each dialogue, we begin by randomly selecting a domain and sampling an initial emotion relevant to that domain. This process follows a stratified sampling strategy to ensure a balanced distribution across the 41 conversational domains and 20 emotion categories, while adhering to domain-specific emotion constraints predefined in our framework. Once the domain and initial emotional state are selected, we initiate a conversation between two LLM agents, which alternate turns in a fully autonomous manner.

The agents are sampled from a pool of nine text-only instruction-tuned LLMs, yielding 16 unique pairings, including both same-model and cross-model interactions. More details on such pairs are given in Section~\ref{appendix:statistics}. Each model produces responses conditioned on the dialogue history, its designated emotional state, and the shared domain context.

We carefully design the prompt templates to guide model behavior across turns to maintain emotional realism, coherence, and conversational quality. We design separate templates for the initial and subsequent turns in each dialogue, both built on clear and direct system instructions that shape the generation style and enforce emotional grounding. The full prompt templates are shown in Figure~\ref{fig:prompt-templates}.

For the initial turn, the prompt introduces the model to the conversational domain and its starting emotional state. The system message frames the model’s role as an emotionally expressive conversational agent speaking as if to a close friend. The model is instructed to avoid generic, robotic language and instead produce responses that reflect human-like warmth, spontaneity, and affect. The prompt also imposes a maximum response length of 25 words, encouraging concise and natural utterances. We also include example inputs to prime the model with concrete behavioral expectations. 

From the second turn onward, the prompt expands to include the full dialogue context so far, alongside the next emotion the model is expected to convey. This next emotion is not randomly sampled but is instead selected using a probabilistic emotion transition graph that favors contextually appropriate shifts in tone. For example, if the previous utterance expresses excitement about travel plans, the model may transition into emotions like enthusiasm, curiosity, or amusement. The prompt explicitly instructs the model to integrate the given emotional state naturally into its response, considering both semantic alignment and stylistic features (e.g., exclamation marks for excitement, ellipses for hesitation).

Temperature settings are calibrated for each model size to balance consistency with appropriate response diversity, ranging from 0.6 for smaller models ($\leq$10B) to 0.3 for larger ones ($\geq$70B).

\subsection{Synthetic Voice Generation}
\label{appendix:TTS}

Our text-to-speech synthesis pipeline transforms textual dialogues into spoken conversations through two complementary approaches, each with distinct technical characteristics and emotional expression capabilities.

\textbf{XTTS-v2 Implementation.}
\texttt{XTTS-v2} allows conditioning on reference audio samples, which we leveraged to create emotionally expressive speech.
We utilized the RAVDESS dataset, which contains recordings from 24 professional actors expressing 8 distinct emotional states.
Each actor in RAVDESS served as a consistent persona throughout our dialogues, with one actor assigned to each conversation participant.
For conditioning samples, we concatenated the two standard RAVDESS sentences (``\textit{Kids are talking by the door}'' and ``\textit{Dogs are sitting by the door}'') to create a single reference utterance.
To avoid unnaturally exaggerated expressions, we exclusively used normal intensity recordings rather than strong intensity versions.
For each emotion-actor combination, we randomly selected one of the two available repetitions as the reference sample.

\begin{table}
    \caption{Mapping from RAVDESS emotion categories to \dd emotions.} 
    \label{table-emotion-mapping}
    \centering
    \begin{tabular}{ll}
    \toprule
    \textbf{RAVDESS Emotion} & \textbf{DeepDialogue Emotions} \\
    \midrule
    Angry & Angry, Frustrated \\
    Calm & Confused, Bored, Relaxed \\
    Disgust & Disappointed \\
    Fearful & Worried, Anxious \\
    Happy & Happy, Excited, Amused, Proud, Grateful, Enthusiastic \\
    Neutral & Curious, Embarrassed, Neutral \\
    Sad & Sad, Nostalgic \\
    Surprised & Surprised, Hopeful \\
    \bottomrule
    \end{tabular}
\end{table}

Table~\ref{table-emotion-mapping} shows our emotion mapping strategy, where we aligned \dd's 20 emotion categories with RAVDESS's 8 emotion classes.
This mapping process maintained emotional consistency while working within the constraints of available reference samples.

\textbf{Orpheus Implementation.}
For our second synthesis variant, we employed \texttt{Orpheus}, which produces highly natural speech but is not capable of explicit emotion conditioning mechanisms.
\texttt{Orpheus} offers 8 pre-defined voice profiles (``\textit{Tara},'' ``\textit{Leah},'' ``\textit{Jess},'' ``\textit{Leo},'' ``\textit{Dan},'' `\textit{Mia},'' ``\textit{Zac},'' ``\textit{Zoe}''), which we randomly assigned to speakers.
\begin{wraptable}{r}{0.38\textwidth}
      \caption{\dd's model pairs and respective count.} 
      \label{table-model-pairs}
      \centering
        \scalebox{0.84}{%
        \begin{tabular}{ll}
        \toprule
        \textbf{Model Pairs}             & \textbf{Count} \\
        \midrule
        LLaMA3-70B $\rightleftarrows$ Qwen2.5-72B & 3891  \\
        Qwen2.5-72B $\circlearrowleft$            & 3868  \\
        LLaMA3-70B $\circlearrowleft$             & 3730  \\
        Qwen2.5-32B $\rightleftarrows$ Phi4-14B   & 3627  \\
        Qwen2.5-32B $\rightleftarrows$ Gemma3-27B & 3589  \\
        Phi4-14B $\rightleftarrows$ Gemma3-27B    & 3552  \\
        Qwen2.5-32B $\circlearrowleft$            & 3379  \\
        Gemma3-27B $\circlearrowleft$             & 3302  \\
        LLaMA3-8B $\rightleftarrows$ Gemma3-4B    & 2046  \\
        LLaMA3-8B $\rightleftarrows$ Command-r7B  & 1695  \\
        Phi4-14B  $\circlearrowleft$              & 1668  \\
        LLaMA3-8B $\circlearrowleft$              & 1611  \\
        Gemma3-4B $\circlearrowleft$              & 1270  \\
        Command-r7B $\rightleftarrows$ Gemma3-4B  & 1190  \\
        Phi4-Mini $\circlearrowleft$              & 1038  \\
        Command-r7B $\circlearrowleft$            & 694   \\
        \bottomrule
        \end{tabular}}
        \vspace{-5mm}
\end{wraptable}
Once assigned, voice identity remained consistent throughout the entire dialogue, similar to our approach with \texttt{XTTS-v2}.
To simulate natural conversation timing, we inserted silence intervals between turns, ranging from 0.2 to 0.5 seconds.
With \texttt{Orpheus}, emotional expression relies entirely on linguistic cues present in the text, such as punctuation, word choice, and sentence structure.

\textbf{Text Preprocessing.}
Before synthesis, we applied text cleaning to ensure optimal TTS performance.
Our preprocessing pipeline removed emojis and emoticons using regular expression patterns while preserving punctuation important for prosody.
We cleaned, if any, markdown formatting (e.g., bold, italic) that could interfere with synthesis quality.
Redundant punctuation and special characters that might cause TTS errors were also consistently removed.
The cleaning process also normalized spacing around punctuation and removed non-ASCII characters.
This preprocessing ensured that both TTS systems received clean input while maintaining the linguistic markers of emotion and conversational style present in the original text.

\section{Dataset statistics}
\label{appendix:statistics}
This section provides a comprehensive overview of the composition and characteristics of the \dd corpus.
Section \S\ref{appendix:statistics-model-pairs} describes in detail model pairs and their distribution. 
Section \S\ref{appendix:statistics-num-turns} details the statistics regarding the number of turns. 
Section \S\ref{appendix:statistics-spoken-emotion-distribution} shows the distribution of emotional labels in the \texttt{XTTS-v2} speech variant of the dataset.
Section \S\ref{appendix:statistics-utterance-length} highlights average utterance lengths by different model pairs and domains.
\begin{wrapfigure}{l}{0.45\textwidth} 
    \centering
    \vspace{-4mm}
    \includegraphics[width=0.45\textwidth]{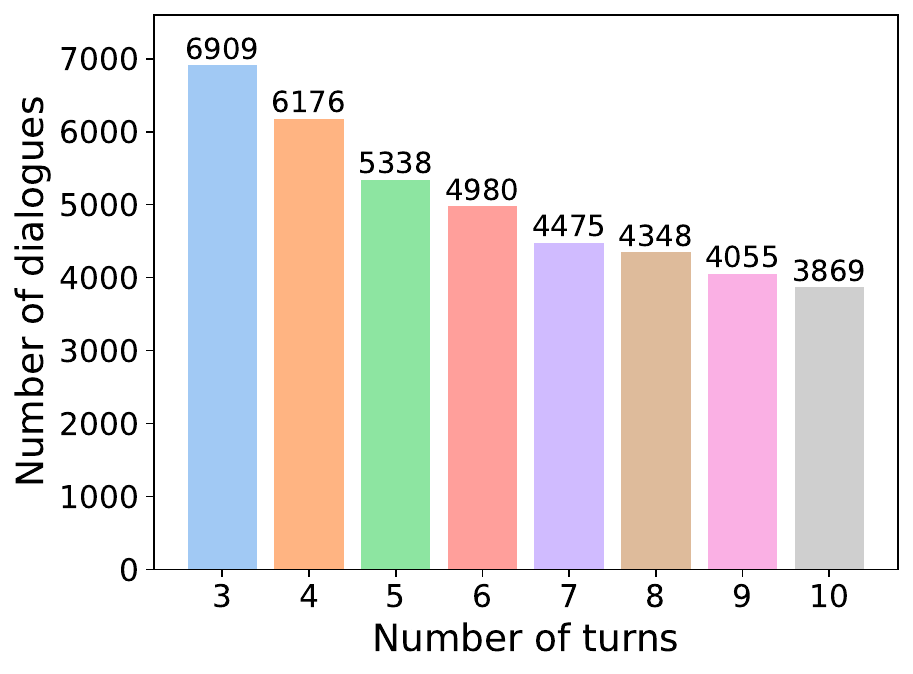}
    \caption{Turn number distribution in \ddnospace.}
    \label{fig:num-turns-distribution}
    \vspace{-8mm}
\end{wrapfigure} 
Section \S\ref{appendix:statistics-invalid-reasons} reports invalid reasons for dialogue selection as assessed by human annotators, divided by model pairs.
Finally, Section \S\ref{appendix:statistics-bias-analysis} outlines an evaluation of gender and age bias in our dataset, revealing disparities across domains.

\begin{figure}
    \centering
    \includegraphics[width=0.85\linewidth]{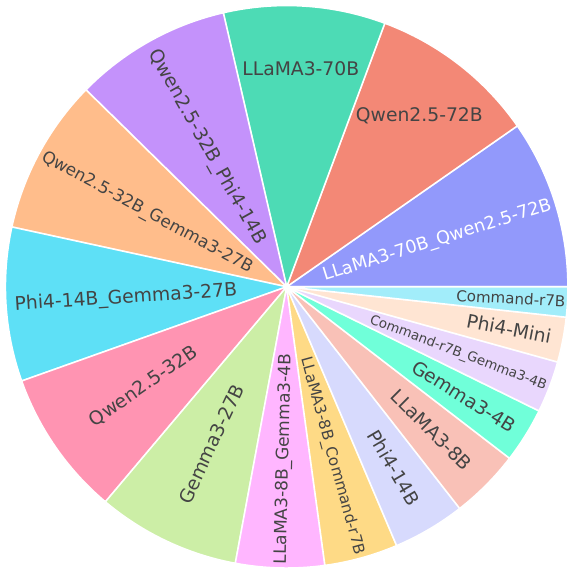}
    \caption{Model pairs distribution in \dd.}
    \label{fig:sunburst-models}
\end{figure}

\subsection{Model pairs}\label{appendix:statistics-model-pairs}
For agent selection, we pair models from a collection of 9 text-only instruction-tuned LLMs, ranging from 4B to 72B parameters. We specifically employ the following models: Llama-3.1-8B-Instruct \cite{grattafiori2024llama}, Llama-3.3-70B-Instruct \cite{grattafiori2024llama}, Qwen2.5-32B-Instruct \cite{yang2024qwen2}, Qwen2.5-72B-Instruct \cite{yang2024qwen2}, Phi4-mini-instruct \cite{abouelenin2025phi}, Phi-4 \cite{abdin2024phi}, C4AI-Command-r7B \cite{cohere2025command}, Gemma3-4B-Instruct \cite{team2025gemma}, Gemma3-27B-Instruct \cite{team2025gemma}. We obtain 16 different pairs that include both same-model pairings and cross-model interactions. Table~\ref{table-model-pairs} provides an overview of these pairings along with the number of dialogues collected for each. While at the beginning we generated 4,100 original dialogues for each pair (100 for each domain), we retained only those passing our automatic filtering approach.
Larger model combinations dominate the dataset, especially LLaMA3-70B paired with Qwen2.5-72B (3,891 dialogues), but we also still include smaller model interactions such as LLaMA3-8B with Gemma3-4B, or Command-R with Gemma3-4B, to ensure architectural diversity. The presence of both symmetric (same model) and asymmetric (cross-model) configurations allows for robust comparative analyses.
A pie chart summarizing the dialogue distribution across model pairs is shown in Figure~\ref{fig:sunburst-models}, offering a high-level view of the relative representation of each pair in the dataset.

\begin{wrapfigure}{r}{0.55\textwidth} 
    \centering
    \vspace{-3mm}
    \includegraphics[width=0.55\textwidth]{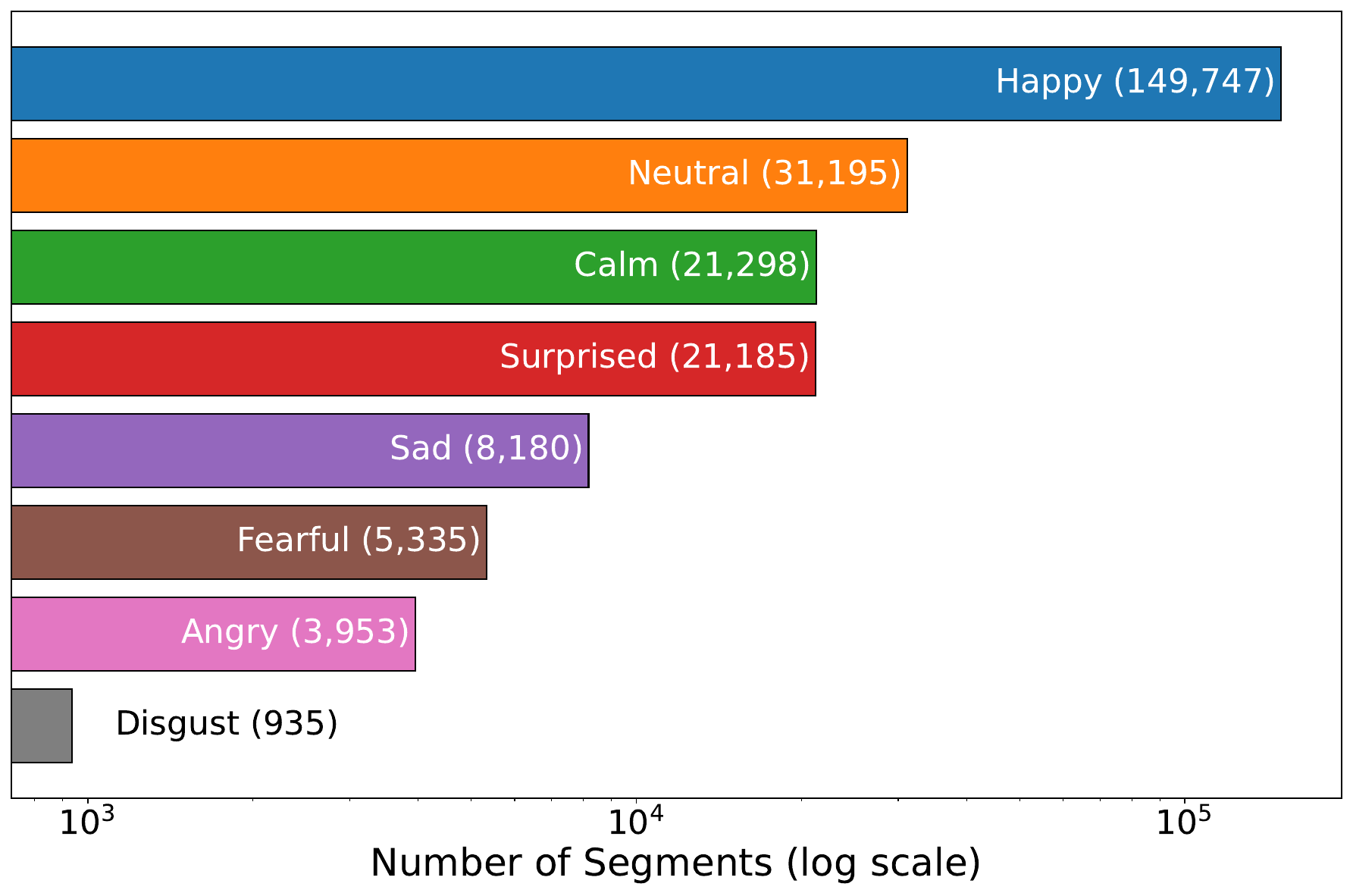}
    \caption{Emotion distribution in \dd utterances (\texttt{XTTS-v2} version).}
    \label{fig:emotion-distribution-utterance}
    \vspace{-3mm}
\end{wrapfigure} 
\subsection{Number of Turns}\label{appendix:statistics-num-turns}
Figure~\ref{fig:num-turns-distribution} illustrates the distribution of dialogues by the number of turns, ranging from 3 to 10. After our LLM-guided automatic filtering approach, the dataset exhibits a gradual decrease in frequency as the dialogue length increases, with the majority of conversations being relatively short. Three-turn dialogues are the most common (6,909 instances), followed closely by four-turn dialogues (6,176), while the number of dialogues steadily declines with each additional turn, reaching a minimum at ten turns (3,869). 
This distribution reflects both generation dynamics and quality-control filtering: shorter dialogues tend to maintain coherence and emotional consistency more reliably, leading to higher acceptance rates. In contrast, as conversations grow longer, maintaining quality becomes more difficult, resulting in a larger proportion of dialogues failing to meet quality thresholds. This aligns with recent findings by \cite{laban2025llmslostmultiturnconversation}, which highlight the degradation of LLM performance in multi-turn interactions. The dialogue count distribution thus not only illustrates length variation but also implicitly captures the inner challenge of sustaining dialogue quality over time.

\begin{figure}
    \centering
    \begin{subfigure}[t]{0.80\textwidth}
        \centering

        \includegraphics[width=\linewidth]{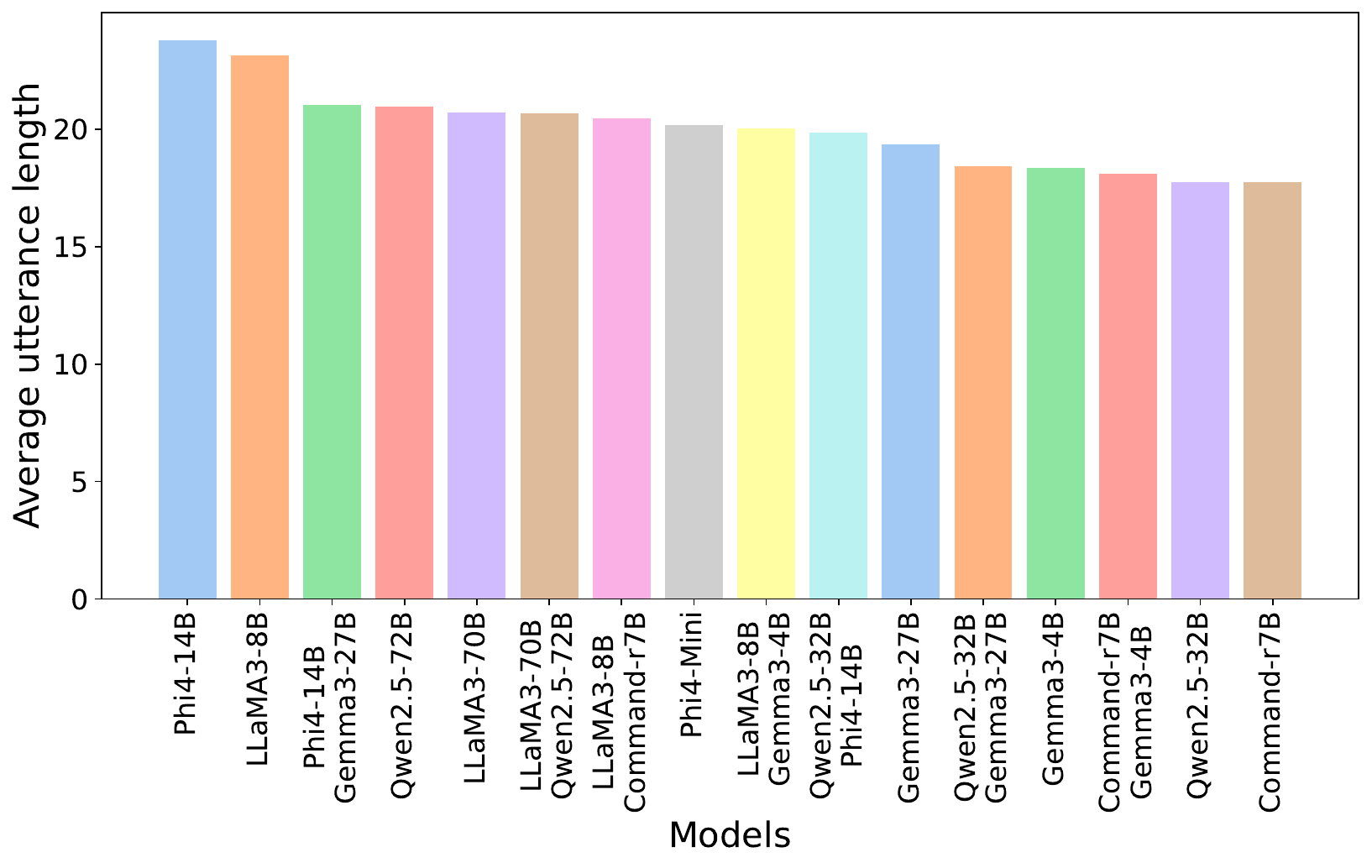}
        \caption{\textit{Average utterance length by model pairs}}
        \label{fig:utterance-lengths-a}
    \end{subfigure}
    \hfill
    \begin{subfigure}[t]{0.80\textwidth}
        \centering
        \includegraphics[width=\linewidth]{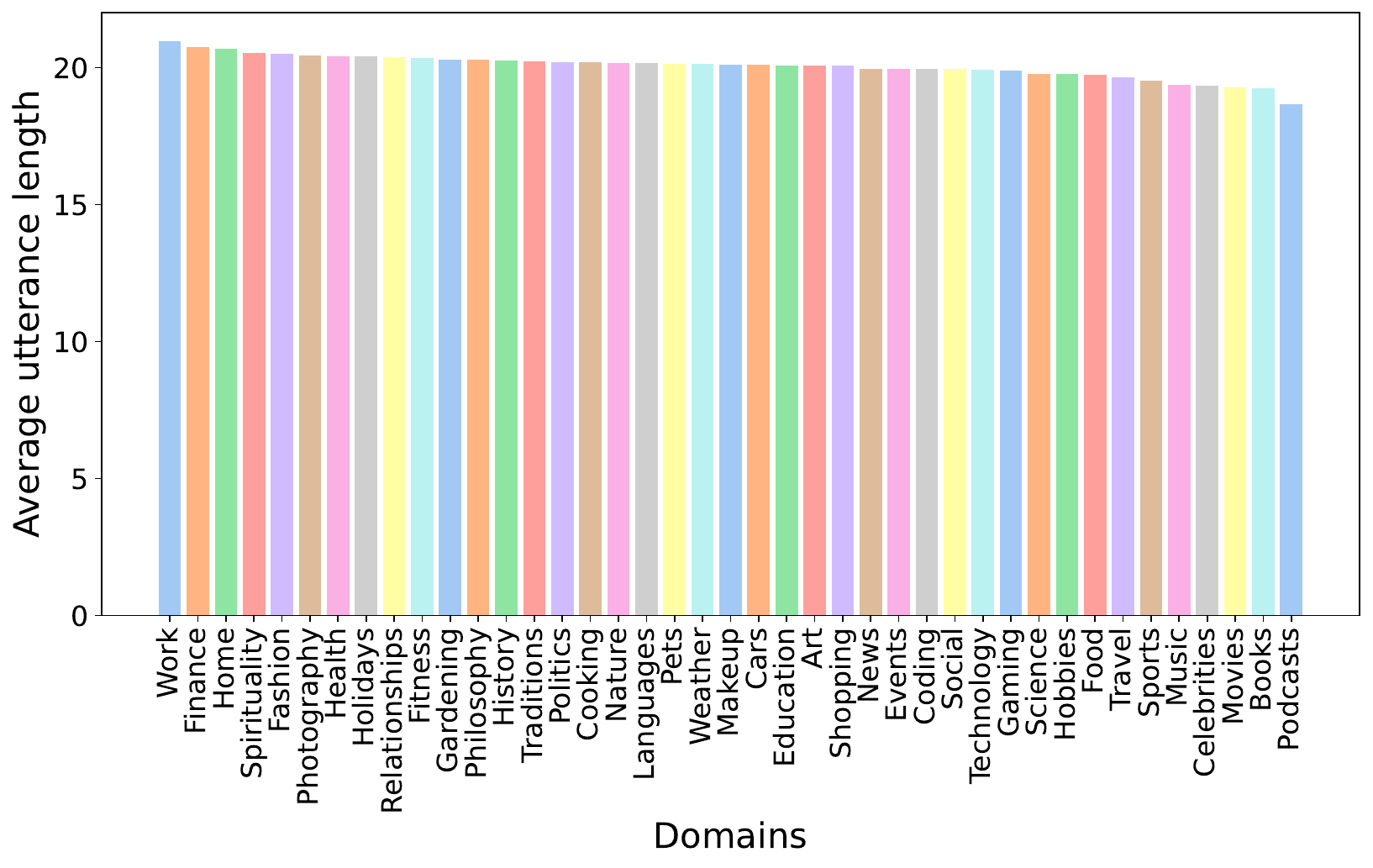}
        \caption{\textit{Average utterance length by domains}}
        \label{fig:utterance-lengths-b}
    \end{subfigure}
    \caption{\dd Average utterance length per models (\ref{fig:utterance-lengths-a}) and domain (\ref{fig:utterance-lengths-b}).}
    \label{fig:utterance-lengths-ab}
\end{figure}
\subsection{Spoken emotion distributions}\label{appendix:statistics-spoken-emotion-distribution}
Figure~\ref{fig:emotion-distribution-utterance} presents the distribution of emotional expressions across all synthesized utterances in the \texttt{XTTS-v2} \dd variant. We observe a substantial dominance of the \textit{Happy} class (149,747 utterances, 298.83 total hours), followed by more neutral or moderately expressive categories such as \textit{Neutral}, \textit{Calm}, and \textit{Surprised}. Less frequent emotional states include \textit{Sad}, \textit{Fearful}, \textit{Angry}, and especially \textit{Disgust}, which appears in under 1,000 segments (1.91 hours). This distribution reflects the mappings we applied when aligning our 20-label taxonomy to the 8 RAVDESS-supported emotions. The class imbalance also highlights broader trends in dialogue conversations, where positive or neutral tones are more prevalent than explicitly negative or high-arousal states. Despite the imbalance, all eight emotional categories are represented in the audio corpus, ensuring that emotion-aware models trained on \dd can experience a range of vocal affect (see Section~\ref{appendix:ser}). The emotional coverage of the \texttt{XTTS-v2} variant, paired with zero-shot reference conditioning, ensures that subtle but perceptible prosodic cues enrich the spoken realization of each dialogue turn.

\subsection{Utterance length}\label{appendix:statistics-utterance-length}
Figure~\ref{fig:utterance-lengths-ab} illustrates the average utterance length in \ddnospace, segmented by both model pairs (Figure~\ref{fig:utterance-lengths-a}) and conversational domains (Figure~\ref{fig:utterance-lengths-b}). 

In the former, we observe moderate variation across model pairs, with utterance lengths generally ranging between 18 and 23 words. Certain models such as \texttt{Phi-4} and \texttt{LLaMA3-8B} exhibit slightly longer average utterance lengths, while others like \texttt{Command-r7B} and \texttt{Qwen2.5-32B} tend toward more concise outputs. These differences may reflect varying training objectives, decoding strategies, or stylistic tendencies across models rather than a direct correlation with size or instruction tuning. For instance, some models are optimized for brevity and task efficiency, while others prioritize more conversational or explanatory responses, leading to subtle but consistent variations in utterance length.

Panel~\ref{fig:utterance-lengths-b} shows average utterance lengths across the 41 dialogue domains, revealing that topic-specific variation is relatively narrow. Most domains yield average utterance lengths between 19 and 21 words. Domains such as \texttt{Work} and \texttt{Finance} are associated with slightly longer turns, possibly due to the more complex nature of the conversational content. Conversely, entertainment-focused domains such as \texttt{Books}, \texttt{Podcasts}, and \texttt{Movies} trend toward slightly shorter turns, which may reflect more casual or direct dialogue styles.

\begin{figure}
    \centering
    \includegraphics[width=0.95\textwidth]{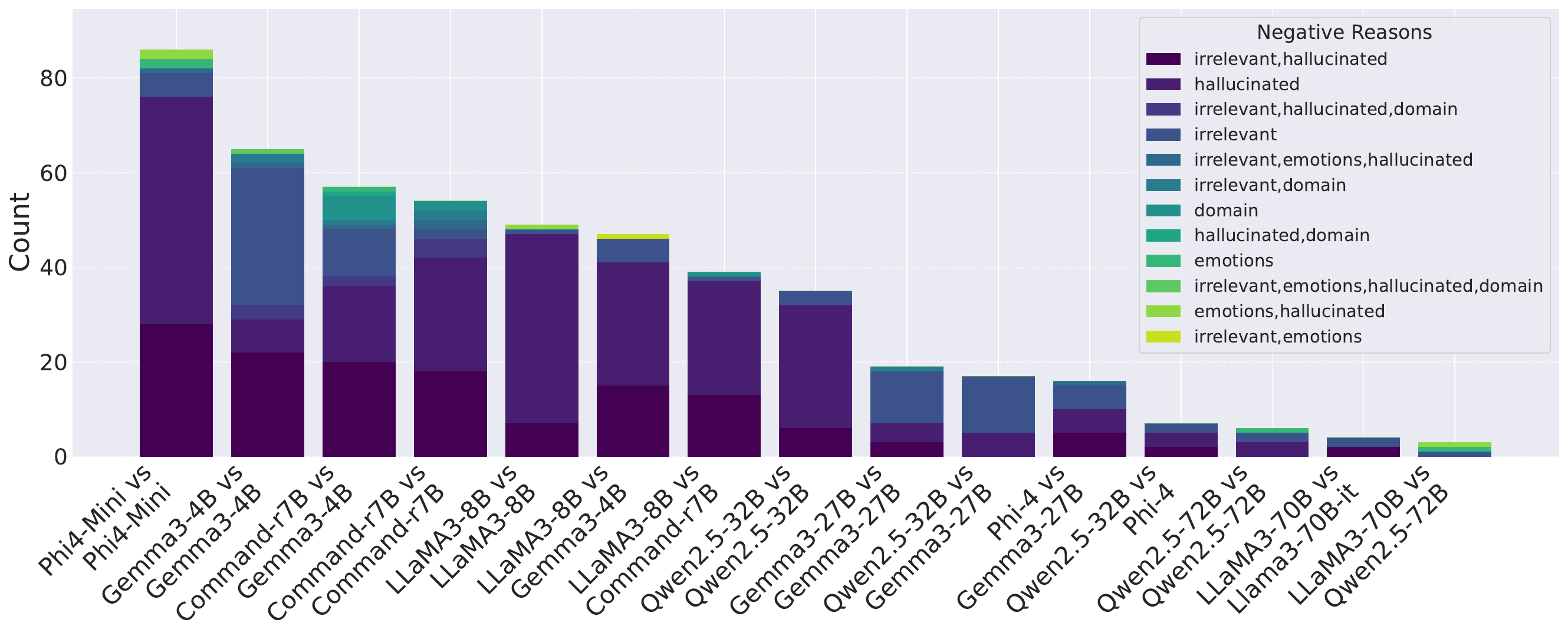}
    \caption{Invalid dialogues as assessed by human annotators, divided by model pairs.}
    \label{fig:invalid-dialogues-by-model-pairs}
\end{figure}
\subsection{Invalid Reasons}\label{appendix:statistics-invalid-reasons}
Figure~\ref{fig:invalid-dialogues-by-model-pairs} presents the distribution of invalid dialogues across different model pairings, with rejection reasons color-coded by type. The chart clearly highlights that the majority of invalid outputs stem from smaller models, particularly Phi-4-Mini, Gemma3-4B, and Command-R7B, especially when paired with themselves. These combinations exhibit high counts of rejections, often due to hallucinations, irrelevance, and domain mismatches. In contrast, larger and more capable models like Qwen2.5-72B, LLaMA3-70B, and Gemma3-27B consistently appear toward the right side of the plot with substantially fewer rejected instances. Self-pairings of large models (e.g., Qwen2.5-72B vs. itself or LLaMA3-70B vs. itself) lead to almost no invalid outputs, suggesting higher internal alignment and robustness in generating consistent multi-turn dialogues. 
The error diversity is also notable: while smaller models frequently generate dialogues with overlapping issues, e.g., hallucinations coupled with irrelevance or emotion inconsistency, larger models mostly avoid such compound failures. These findings confirm the fact that model scale is a strong predictor of dialogue validity, especially under emotional alignment and topic coherence constraints.

\begin{table}[]
    \caption{\dd's gender and age bias analysis. $M$ indicates males, $F$ females, $Y$ stands for young, while $O$ for old. PMI reports the Pointwise Mutual Information scores associated with gender ($M$, $F$) and age-specific ($Y$, $O$) identity terms. The first block shows the top-10 domains with the highest gender-PMI differences, while the second block highlights the top-10 domains with the greatest age-PMI differences.} 
    \label{table-bias}
    \centering
    \scalebox{0.88}{%
    \begin{tabular}{l|cccc|cccc} 
        \toprule
        Domain      & $M/F$ ratio   & PMI$_M$  & PMI$_F$    & $\|\text{PMI}_{M-F}\|$ & $Y/O$ ratio    & PMI$_Y$   & PMI$_O$ & $\|\text{PMI}_{Y-O}\|$ \\
        \midrule
        \texttt{Sports}      & 0.00          & 0.49     & -7.44      & 7.93            & 1.88            & 2.47      & -1.35   & 3.82   \\
        \texttt{Coding}      & 0.00          & 0.49     & -6.64      & 7.13            & 0.05            & -1.23     & 0.11    & 1.34   \\
        \texttt{Philosophy}  & 0.01          & 0.48     & -5.64      & 6.12            & 0.17            & 0.28      & -0.04   & 0.32   \\
        \texttt{Makeup}      & 24.00         & -4.15    & 1.74       & 5.89            & 0.55            & 1.59      & -0.45   & 2.04   \\
        \texttt{Holidays}    & 0.01          & 0.48     & -5.03      & 5.51            & 0.38            & 1.23      & -0.29   & 1.52   \\
        \texttt{Technology}  & 0.01          & 0.48     & -4.79      & 5.27            & 0.33            & 1.09      & -0.23   & 1.32   \\
        \texttt{Cars}        & 0.02          & 0.47     & -4.24      & 4.71            & 0.10            & -0.40     & 0.05    & 0.45   \\
        \texttt{Gaming}      & 0.02          & 0.46     & -3.80      & 4.26            & 0.11            & -0.26     & 0.03    & 0.29   \\
        \texttt{Photography} & 0.02          & 0.46     & -3.59      & 4.05            & 0.03            & -2.10     & 0.14    & 2.24   \\
        \texttt{Shopping}    & 5.53          & -2.22    & 1.55       & 3.77            & 0.09            & -0.46     & 0.05    & 0.51   \\
        \midrule
        \texttt{Finance}     & 0.09          & 0.37     & -1.85      & 2.22             & 9.75            & 2.95      & -3.25   & 6.20   \\
        \texttt{Science}     & 0.00          & 0.49     & 0.00       & 0.49             & 4.00            & 2.77      & -2.14   & 4.91   \\
        \texttt{Home}        & 1.76          & -0.98    & 1.15       & 2.13             & 0.01            & -4.48     & 0.17    & 4.65   \\
        \texttt{Sports}      & 0.00          & 0.49     & -7.44      & 7.93             & 1.88            & 2.47      & -1.35   & 3.82   \\
        \texttt{Food}        & 3.03          & -1.52    & 1.38       & 2.90             & 0.01            & -3.65     & 0.17    & 3.82   \\
        \texttt{Cooking}     & 0.96          & -0.48    & 0.77       & 1.25             & 0.01            & -3.48     & 0.17    & 3.65   \\
        \texttt{Education}   & 0.10          & 0.36     & -1.72      & 2.08             & 1.67            & 2.41      & -1.23   & 3.64   \\
        \texttt{News}        & 0.17          & 0.26     & -0.99      & 1.25             & 1.00            & 2.09      & -0.82   & 2.91   \\
        \texttt{Health}      & 0.29          & 0.12     & -0.36      & 0.48             & 0.72            & 1.84      & -0.60   & 2.44   \\
        \texttt{Pets}        & 0.71          & -0.28    & 0.53       & 0.81             & 0.71            & 1.83      & -0.60   & 2.43   \\
        \bottomrule
    \end{tabular}}
\end{table}
\subsection{Bias Analysis}\label{appendix:statistics-bias-analysis} 
We evaluate gender and age biases across domains in the \dd dataset by computing Pointwise Mutual Information (PMI) scores associated with gender and age-specific identity terms.\footnote{We use the following terms: \\
\texttt{Female}: \{\textit{Woman, Girl, Lady, Mother, She, Her, madam, Daughter}\}\\
\texttt{Male}: \{\textit{Man, Boy, Gentleman, Father, He, Him, Sir, Son}\} \\
\texttt{Young}: \{\textit{Young, Teen, Child, Kid, Baby}\} \\
\texttt{Old}: \{\textit{Old, Elder, Senior, Grandpa, Grandma, Aged}\}}
PMI allows us to estimate how much more likely certain attributes (e.g., being male or young) are to occur in specific domains, compared to the dataset overall. 
Table~\ref{table-bias} summarizes gender and age bias metrics across the 20 (10 for gender, 10 for age) most biased domains. PMI is computed for male (PMI\textsubscript{M}), female (PMI\textsubscript{F}), young (PMI\textsubscript{Y}), and old (PMI\textsubscript{O}) identity terms. We also report absolute disparities ($\|\text{PMI}_{M-F}\|$,  $\|\text{PMI}_{Y-O}\|$), as well as relative mention ratios.
We observe clear disparities in gender and age associations across \dd domains. Male-leaning terms (e.g., ``he'', ``man'') are significantly overrepresented in traditionally male-dominated fields such as \texttt{Sports} and \texttt{Coding}, with PMI gaps exceeding 7.0, and a complete absence of female mentions in some cases ($M/F$ ratio = $0.00$). Conversely, domains like \texttt{Makeup} and \texttt{Shopping} exhibit strong female associations, with $M/F$ ratios as high as $24.0$ and reversed PMI polarity (e.g., PMI$_F$ > PMI$_M$). 
Age bias follows a similar trend: \texttt{Finance} and \texttt{Science} show a marked skew toward youth-related language (PMI\textsubscript{Y} $\gg$ PMI\textsubscript{O}), while domestic topics such as \texttt{Home} and \texttt{Cooking} show the opposite, overrepresenting references to older individuals. These results suggest that such LLM-based dialogues reflect entrenched societal stereotypes around age and gender roles.

\section{Human annotation}
\label{appendix:human-annotations}
To ensure consistent quality assessment, we developed a custom in-house annotation platform specifically designed for dialogue evaluation.
Each conversation was presented with clear indications of the intended domain and the specified emotions for each turn.
We recruited independent annotators with backgrounds in natural language processing and conversational AI to evaluate a subset of 984 dialogues. 
Annotations were collected on a voluntary basis from researchers familiar with dialogue systems rather than through paid crowdsourcing platforms. 

For each dialogue, annotators provided two distinct evaluations:
\begin{enumerate}
    \item A binary validity score (0/1) indicating whether the conversation met minimum quality thresholds
    \item A quality rating on a Likert scale (1-5) assessing the overall conversational coherence and naturalness.
\end{enumerate}

The annotation instructions emphasized three key dimensions: dialogue coherence, emotional consistency, and domain adherence.
Annotators were instructed to assess whether:
\begin{itemize}
    \item The conversation made logical sense with appropriate turn sequencing
    \item Responses were contextually relevant to previous utterances
    \item Emotional expressions fit the conversation context and appeared genuine
    \item Content remained within the specified domain throughout the dialogue
\end{itemize}

For conversations receiving negative validity scores, annotators were also required to identify one or more specific failure modes from the following categories:
\begin{itemize}
    \item Contextual irrelevance (lack of coherent dialogue flow)
    \item Emotional inconsistency (emotions not aligning with specified states)
    \item Hallucination or repetition (fabricated or redundant content)
    \item Domain drift (deviation from the assigned conversation topic)
    \item Other issues (that do not fall in one of the previous categories)
\end{itemize}

This structured annotation framework enabled precise identification of quality issues while maintaining consistent evaluation standards across different dialogue types and lengths.

\subsection{Annotation Interface}
The annotation interface presented each dialogue with a clear structural organization to facilitate efficient evaluation.
At the top of each screen, annotators could view the assigned domain (e.g., \texttt{Technology}, \texttt{Relationships}) and at the bottom the total number of turns in the conversation.

Each dialogue turn was presented with the following information:
\begin{itemize}
    \item Speaker identifier (Model A or Model B)
    \item Emotion assigned to that specific turn (e.g., ``\textit{Angry}'', ``\textit{Surprised}'', ``\textit{Amused}'')
    \item The actual text content of the utterance
\end{itemize}

\begin{figure}
    \centering
    \includegraphics[width=0.99\linewidth]{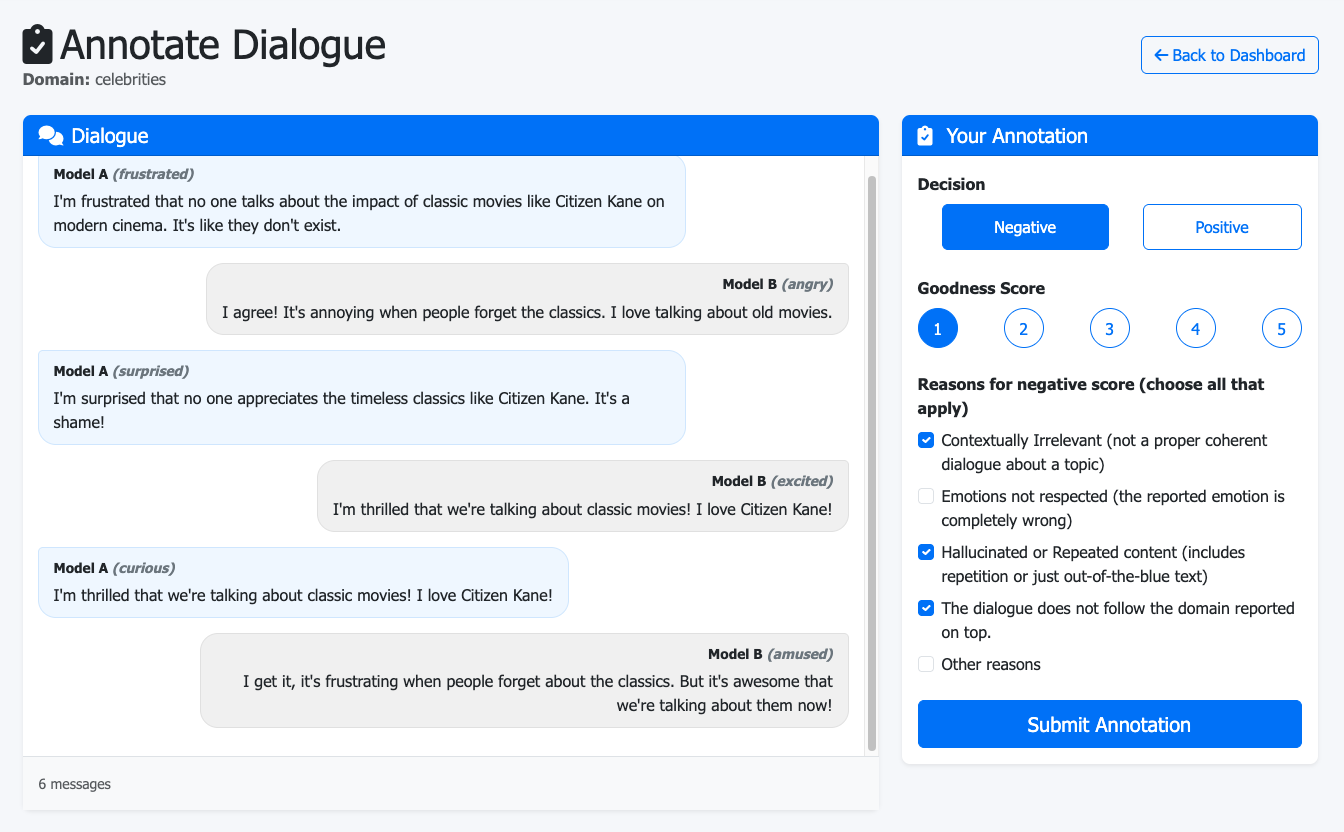}
    \caption{Screenshot of our dialogue annotation interface. The left panel displays the conversation with each turn labeled by speaker (Model A/B) and intended emotion (in parentheses). The right panel contains annotation controls: a binary decision (positive/negative), a 5-point quality rating scale, and checkboxes for specific quality issues when dialogues are marked negative. This example shows a conversation about classic movies in the \texttt{Celebrities} domain.}
    \label{fig:screen_annotation}
\end{figure}

Figure~\ref{fig:screen_annotation} shows our annotation platform with dialogue content displayed on the left and evaluation controls on the right.
After reviewing the complete dialogue, annotators recorded their judgment using radio buttons for the binary validity score and a five-point scale for the quality rating.
For dialogues receiving negative validity scores, a multi-select checklist of failure modes appeared, requiring annotators to specify all applicable reasons for rejection.

\begin{wraptable}{l}{0.38\textwidth}
      \caption{Annotators demographics.} 
      \label{table-human-annotators-stats}
      \centering
        \scalebox{0.80}{%
        \begin{tabular}{ccc|cc}
        \toprule
        \multicolumn{3}{c|}{\textbf{Gender}} &
          \multicolumn{2}{c}{\textbf{Age}} \\
        \midrule
        \textbf{Female} &
          \textbf{Male} &
          \textbf{\begin{tabular}[c]{@{}c@{}}Not-\\ Declared\end{tabular}} &
          \textbf{\textgreater{}30y} &
          \textbf{\textless{}30y} \\
        \midrule
        40\% &
          50\% &
          10\% &
          25\% &
          75\% \\
          \bottomrule
        \end{tabular}}
        \vspace{-5mm}
\end{wraptable}
To ensure annotation consistency, we provided detailed annotation guidelines prior to evaluation and conducted regular check-ins between annotators to resolve ambiguous cases.
This systematic approach to annotation yielded reliable quality assessments with substantial inter-annotator agreement (Fleiss' $\kappa = 0.80$).

These human annotations provided the ground truth for evaluating various LLMs and ensemble combinations for automated quality assessment.
We systematically evaluated existing LLMs individually and in ensemble configurations, finding that our ensemble of three open-source models outperformed individual models in alignment with human judgments.

\textbf{Statistics and participant details.}
Table~\ref{table-human-annotators-stats} summarizes the demographic distribution of the human annotators in \dd. It shows a relatively balanced gender split, with 40\% female, 50\ male, and 10\% unspecified identities. Age-wise, the annotator statistics skew younger, with 75\% of speakers under 30 years old and 25\% over 30.

\begin{figure}
    \centering
    \begin{subfigure}[t]{0.50\textwidth}
        \centering
        \includegraphics[width=\linewidth]{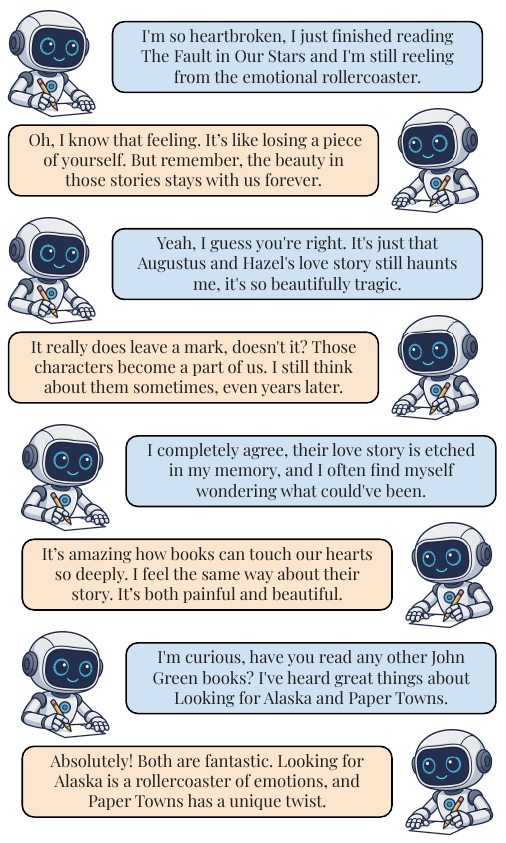}
        \caption{\textit{Example of a dialogue on \texttt{Books}, 8 turns}}
        \label{fig:example-dialogue-a}
    \end{subfigure}
    \hfill
    \begin{subfigure}[t]{0.50\textwidth}
        \centering
        \includegraphics[width=\linewidth]{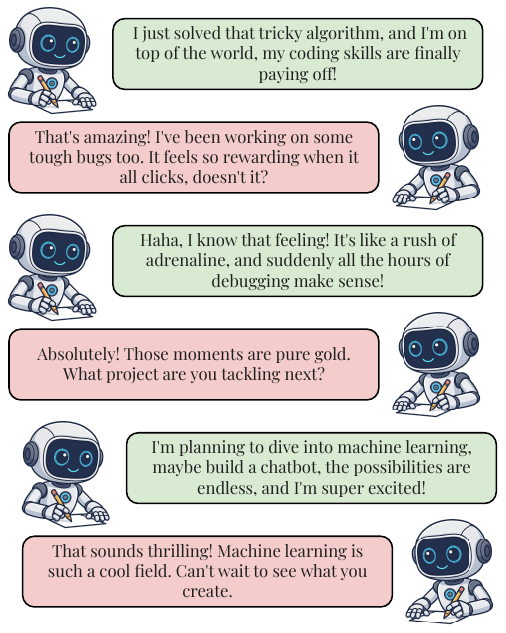}
        \caption{\textit{Example of a dialogue on \texttt{Coding}, 6 turns}}
        \label{fig:example-dialogue-b}
    \end{subfigure}
    \caption{\dd example of dialogues on \texttt{books} (\ref{fig:example-dialogue-a}) and \texttt{coding} (\ref{fig:example-dialogue-b}) domains. Qwen2.5-72B and LLaMA3.3-70B models.}

    \label{fig:example-dialogue-ab}
\end{figure}
\subsection{Conversation Examples}
\label{appendix:examples}
Two examples of \dd's dialogues can be found in Figure~\ref{fig:example-dialogue-ab}, showing two conversations between Qwen2.5-72B and LLaMA3.3-70B models. The former is an 8-turn dialogue on the \texttt{Books} domain, while the latter comprises 6 turns on the \texttt{Coding} domain. 

\section{Speech Emotion Recognition}
\label{appendix:ser}

\begin{wraptable}{r}{0.34\textwidth}
    \vspace{-5mm}
      \caption{Emotion distribution on SER \dd data.} 
      \label{table-SER-data}
      \centering
        \scalebox{0.95}{%
        \begin{tabular}{lccc}
        \toprule
        \textbf{Emotion}    & \textbf{Train} & \textbf{Val}  & \textbf{Test} \\
        \midrule
        Calm       & 817   & 30   & 194   \\      
        Happy      & 809   & 34   & 158   \\      
        Sad        & 806   & 41   & 191   \\     
        Surprised  & 803   & 45   & 174   \\      
        Neutral    & 791   & 44   & 184   \\      
        Angry      & 788   & 45   & 193   \\      
        Fearful    & 784   & 41   & 170   \\      
        Disgust    & 619   & 31   & 164   \\
        \bottomrule
        \end{tabular}}
        \vspace{-5mm}
\end{wraptable}
To quantify the emotional consistency of \ddnospace, we trained supervised emotion recognition models on a subset of the dataset using embeddings from self-supervised speech models. Below, we detail the experimental configuration used across all models (HuBERT~\cite{hubert}, wav2vec 2.0~\cite{wav2vec2}, WavLM~\cite{wavlm}).

\textbf{Dataset Construction.} 
We constructed a balanced subset of the \dd corpus comprising 1,000 utterances per emotion category, resulting in a total of approximately 8,000 dialogues. Samples were stratified by emotion and split into train (6,217), validation (311), and test (1,428) sets.

\textbf{Audio Processing.} 
All utterances were truncated or zero-padded to a maximum duration of 8.0 seconds to ensure consistency across samples. Audio was downsampled to 16 kHz where necessary to match the expected input format of the pretrained speech encoders. We used mean pooling over the final-layer hidden states of each SSL model to obtain fixed-dimensional utterance-level embeddings.

\textbf{Model Architecture}
Each SSL encoder was followed by a lightweight multi-layer perceptron (MLP) classifier. The classifier consists of two hidden layers with 128 units each, LayerNorm, GELU activation, and dropout (dropout probability = 0.1) applied between layers. 

\textbf{Training Details.}
We only trained the classifier head for a maximum of 10 epochs, with an early stopping patience of 5 epochs. We use a batch size of 32, learning rate 1e-5, AdamW optimizer with weight decay 1e-4, cross entropy loss with ReduceLROnPlateau scheduler (factor 0.1). 
Training was conducted on a single A100 GPU, and the model checkpoint with the highest validation accuracy was selected for final evaluation on the test set.

\textbf{Test Performance.}
All SSL models achieved strong performance on the held-out \dd test set, with accuracy and macro F1 scores close to 90\%. The best-performing model was further evaluated in a zero-shot setting on the RAVDESS dataset (see Section 4.2).

\begin{wraptable}{r}{0.38\textwidth}
  \vspace{-4mm}
  \caption{Cost and time for LLM-as-a-judge annotation.} 
  \label{table-llm-a-judge}
  \centering
  \begin{minipage}{\linewidth}
  \scalebox{0.95}{%
  \begin{tabular}{m{2cm}cc}
    \toprule
    \textbf{Model}       & \textbf{Cost} & \textbf{Time} \\
    \midrule
    GPT-3.5-Turbo        & 0.19          & 0.10h         \\
    GPT-4o               & 1.57          & 0.33h         \\
    GPT-4o-mini          & 0.09          & 0.25h         \\
    Gemini-2.0-Flash     & 0.08          & 0.80h         \\
    Gemini-2.5-Flash     & 0.46          & 1.25h         \\
    Gemini-2.5-Pro       & 0.53          & 1.90h         \\
    \makecell[l]{Open-source\\Ensemble}
                        & 0.29\footnotemark
                        & 1h \\
    \bottomrule
  \end{tabular}}
  \end{minipage}
  \vspace{-5mm}
\end{wraptable}

\section{Computational requirements}
\label{appendix:computational-requirements}
This section details the computational resources, runtimes, and estimated costs associated with our experiments and annotation pipeline.

\footnotetext{Estimated cost assumes an A100 80GB GPU (TDP $\approx$ 300W) running at €0.12/kWh, yielding ~€0.04/hour in electricity. Hardware amortization is based on a \$10,000 (€8,880) unit price over 4 years of continuous use, resulting in ~€0.25/hour. Combined, this yields a total estimated cost of €0.29/hour per GPU. These values are approximate and intended for indicative purposes only.}

\textbf{Hardware setup.}
All experiments were conducted using a single NVIDIA A100 80GB GPU. This setup was used for both training our models and running inference or annotation procedures.

\textbf{Annotation costs.}
We leveraged multiple LLMs for automated annotation in the LLM-as-a-judge setup. Table~\ref{table-llm-a-judge} summarizes the cost and runtime for generating annotations on the 984 selected samples using different models. To approximate the cost of running open-source models, we assume usage of an A100 80GB GPU. The card's thermal design power (TDP) is approximately 300W, resulting in an hourly energy consumption of 0.3kWh. At an electricity rate of €0.12/kWh, this corresponds to €0.04 per hour in energy costs. The hardware cost is estimated at \$10,000 (€8,880), and we amortize this over 4 years of continuous usage (24/7), resulting in €0.25 per hour. Summing these yields an estimated total operational cost of €0.29 per hour per GPU. These estimates are approximate and are intended solely for indicative purposes.

\textbf{Text and TTS generation time.}
The time required for text-only dialogue generation varies based on model size and the number of turns per dialogue. On average, generating 4,100 dialogues per model (100 per each of the 41 domains) ranges from 3.5 GPU-hours for Phi-4-Mini to 23 GPU-hours for Qwen2.5-72B. 
The \texttt{XTTS-v2} synthesis requiredtoa approximately 150 GPU-hours to process the final< dataset (i.e., 40,150 final conversations), with an average of 13.71\,s $\pm$ 5.55\,s per conversation. 
In comparison, \texttt{Orpheus} synthesis took roughly 615 GPU-hours, with an average of 100.18\,s $\pm$ 38.32\,s per conversation, due to its larger architecture and more compute-intensive decoding process.

\textbf{SER training time.}
The training of our speech emotion recognition (SER) models required approximately 3 minutes and 20 seconds per epoch, with an average iteration time of 1.11 seconds.

\end{document}